\documentclass{article}
\usepackage[a4paper]{geometry}
	\usepackage{graphicx}
	\usepackage[pdftex,breaklinks, colorlinks, citecolor=blue, linkcolor=blue,urlcolor=blue]{hyperref}

\usepackage{amsmath}
\usepackage{amssymb}
\usepackage{amsfonts} 
\usepackage{graphicx}
\usepackage{url}
\usepackage[caption=false,font=footnotesize]{subfig}

\newcommand{\ie}{i.e., }
\newcommand{\eg}{e.g., }

\newtheorem{theorem}{Theorem}[section]

\newtheorem{definition}[theorem]{Definition}

\newtheorem{example}[theorem]{Example}

\def \IZ{\mathbb Z}

\def \DD{\mathbb D}
\def \GG{\mathbb G}
\def \HH{\mathbb H}
\def \LL{\mathbb L}
\def \NN{\mathbb N}

\def \RR{\mathbb R}

\def \UU{\mathbb U}

\def \m {$L_\text{min}$}
\def \M {$L_\text{MAX}$}
\def \dist {$D_\text{win}$}

\DeclareMathOperator*{\argmax}{argmax}
\newcommand{\SYSTEM}{{MUDOS-NG}}

\title{\SYSTEM{}: Multi-document Summaries Using N-gram Graphs (Tech Report)}

\author{George Giannakopoulos\\NCSR Demokritos, Greece \and George Vouros\\University of the Aegean, Greece \and Vangelis Karkaletsis\\ NCSR Demokritos, Greece}

\begin{document}
\maketitle
\begin{abstract}
This report describes the \SYSTEM{} summarization system, which applies a set of language-independent and generic methods for generating extractive summaries.
The proposed methods are mostly combinations of simple operators on a generic character n-gram graph representation of texts. 
This work defines the set of used operators upon n-gram graphs and proposes using these operators within the multi-document summarization process in such subtasks as document analysis, salient sentence selection, query expansion and redundancy control. Furthermore, a novel chunking methodology is used, together with a novel way to assign concepts to sentences for query expansion.
The experimental results of the summarization system, performed upon widely used corpora from the Document Understanding and the Text Analysis Conferences, are promising and  
provide evidence for the potential of the generic methods introduced. This work aims to designate core methods exploiting the n-gram graph representation, providing the basis for more advanced summarization systems.
\end{abstract}

\section{Introduction}
Since the late 50s and Luhn~\cite{luhn1958acl} the information community has expressed its interest in summarizing texts. The domains of application of such methodologies range from news summarization~\cite{wu2003obt,barzilay2005sfm,radev2005nso} to scientific article summarization~\cite{teufel2002ssa} and meeting summarization~\cite{niekrasz2005obd,erol2003msm}.

Summarization has been defined as a reductive transformation of a given set of texts, usually described as a three-step process: selection of salient portions of text, aggregation of the information for various selected portions and abstraction of this information, and finally, presentation of the final summary text~\cite{mani1999ssa,jones1999asf}. The summarization community aims to address major problems that arise during the summarization process.
\begin{itemize}
 \item How can one detect and select salient information to be included in the summary? Does the use of a query drive the information-selection task, and how?
 \item How can one assure that the final summary does not contain redundant or repeated information, especially when multiple documents are used as input to the summarization process?
 \item Can one develop methods that will function independently from the language of documents and on what degree can this be achieved?
\end{itemize}

Up to date, many summarization systems have been developed and presented, especially within such endeavors as the Document Understanding Conferences (DUC) and Text Analysis Conferences (TAC)\footnote{See \url{http://duc.nist.gov/} and \url{http://www.nist.gov/tac/}.}. 
The summarization community appears to have moved from single-text to multi-text input and has also reached such domains as opinion summarization and 
``trend'' summarization, as in the case of NTCIR\footnote{See \url{http://research.nii.ac.jp/ntcir/}}. 
However, different evaluations performed in recent years have proved that the multi-summarization task is highly complex and demanding, and that automatic summarizers have a long way to go to perform equally well to humans~\cite{dang2005od,dang2006od,dang2008ous}. 
It was recently shown~\cite{genest2009hextac} that the extractive approach has an upper limit of performance, which is lower when compared to the abstractive approach of humans. However, extractive summarization appears to have more room for improvement in order to reach that upper limit of performance, which is set by humans performing extractive summarization through simple sentence selection and reordering.

Even the current methods for the evaluation of summaries are under criticism~\cite{SparckJones2007ass,conroy2008mtg,ggianna2008tun}. Indeed, it has been shown that evaluating different aspects of a summary is far from being a trivial task. Nevertheless, when it comes to ranking summarization systems, \ie{} the average performance of a summarization system over a set of summaries, the existing evaluation methodologies offer quite good correlation to human judgment~\cite{dang2006od,dang2008ous,ggianna2008sse}.

Within this work we tackle the problems of salience detection and redundancy control in extractive multi-document summarization using a unified, language independent and generic framework based on n-gram graphs. 
The contributed methods offer a basic, language-neutral, easily adaptable set of tools. The basic idea behind this framework is that neighborhood and relative position of characters, words and sentences in documents 
offer more information than that of the `bag-of-words' approach. Furthermore, the methods go beyond the word level of analysis into the sub-word (character n-gram) level, which offers further flexibility and independence from language and acts as a uniform representation for sentences, senses, documents and document sets. Through this study we provide a proof-of-concept methodology that can be used in more advanced summarization systems. We also experiment using this methodology as a basic summarization system, named \SYSTEM{}.

In the following sections we briefly review the current literature (section \ref{sec:Literature}), we present the proposed approach (section \ref{sec:Methodology}) and perform a set of experiments to evaluate its performance and show its potential (section \ref{sec:Experiments}). The article concludes with discussion and proposals for further development (section \ref{sec:Discussion}).

\section{The Literature}\label{sec:Literature}
Presently, the literature of automatic multi-document summarization has grown to a level that is very hard to overview in detail~\cite{SparckJones2007ass}. However, one can identify specific commonalities in the way summarizers extract and reproduce information into output summaries. Summarizing systems are usually classified as being either \textbf{extractive} or \textbf{abstractive} in their approach~\cite{mani1999ssa}. Extractive approaches focus on the extraction and use of text \textbf{chunks}, \ie{} text snippets, from the source texts in the final summary. Abstractive approaches, on the other hand, aim to first represent information using an intermediate representation, for example first-order logic, and then use language generation to produce the output summary from the representation. 
Even though it has been shown that humans summarize in an abstractive fashion~\cite{banko2004ung,endres-niggemeyer-humanstyle}, many current systems continue to use the extractive paradigm to perform summarization. This may be due to the lack of highly robust text-to-intermediate-representation methods as well as natural language generation methods. On the other hand, the description of a summarization system as purely abstractive can be disputed and it appears that summarization systems can differentiate themselves according to their purpose~\cite{SparckJones2007ass}.

In the following paragraphs we overview the approaches existing for salience detection and redundancy removal, as this is the focus of our proposed work. We also make a brief literature review for graph-based methods, to introduce the reader to related work from the domain of graphs.
\subsection{Salience Detection}
To determine salience of information, researchers have used \emph{positional and structural} properties of the judged sentences with respect to the source texts. These properties can be the sentence position (\eg{} number of sentence from the beginning of the text, or from the beginning of the current paragraph) in a document, or the fact that a sentence is part of the title or of the abstract of a document~\cite{edmundson1969nma,radev2004cbs}. 
Also, the \emph{relation} of sentences with respect to a user-specific query or to a specified topic~\cite{conroy2005cqb,varadarajan2006sqs,park2006qbs} are features providing evidence towards importance of information. Cohesion (proper name anaphora, reiteration, synonymy, and hypernymy) and coherence (based on Rhetorical Structure Theory~\cite{mann1987rst}) relations between sentences were used in~\cite{mani1998using} to define salience. Based on a graph representation where each sentence is a vertex and vertices are connected when there is a cohesion or coherence relation between them (\eg{} common anaphora), the salience of a sentence is computed as an operation dependent on the graph representation (\eg{} spreading activation starting from important nodes).


Often, following the \textbf{bag-of-words} assumption, a sentence is represented as a word-feature vector  \eg{}~\cite{torralbo2005dot}. In such cases, the sequence of the represented words is ignored. 
The vector dimensions represent word frequency or the Term Frequency -- Inverse Document Frequency (TF-IDF) value of a given word in the source texts. In other cases, further analysis is performed, aiming to reduce dimensionality and 
produce vectors in a \textbf{latent topic space}~\cite{steinberger2004uls,flores2008bsv}.  Vector representations can be exploited for measuring the semantic similarity between information chunks, by using measures such as the cosine distance or Euclidean distance between vectors.

When the feature vectors for the chunks have been created, clustering of vectors can be performed for identifying clusters corresponding to specific topics.
A cluster can then be represented by a single vector, for example the centroid of the corresponding cluster's vectors~\cite{radev2000cbs}. Chunks closest to these representative vectors are considered to be the most salient. 
We must point out that for the aforementioned vector-based approaches one needs to perform preprocessing to avoid pitfalls due to stop-words and inflection of words that create feature spaces of very high dimension. 
However, the utility of the preprocessing step, which usually involves stemming and stop-word removal, is an issue of dispute~\cite{ledeneva2008eas,leite2007eas}.

More recent approaches use machine learning techniques and sets of different features to determine whether a source text chunk (sentence) should be considered salient and included in the output summary. In that case the feature vector calculated for every sentence may include information like sentence length, sentence absolute position in the text, sentence position within its corresponding paragraph, number of verbs and so forth (\eg{} see~\cite{teufel2002ssa}). 

It has been shown that for specific tasks, like the news summarization task of DUC, simple positional features for the determination of summary sentences can prove very promising for summarization systems~\cite{dang2005od}. However, this may falsely lead to the expectation that features like the ones corresponding to the `first-sentence' heuristic, \ie{} the features that indicate whether a sentence appears to be similar in content and properties to the first sentences of a set of training instances, can be used as a universal rule. The example of short stories is an example where a completely different approach is taken to perform the summarization~\cite{kazantseva2010summarizing}: The summary is expected to describe the setting without giving away the plot. In~\cite{jatowt2006temporal}, we find an approach where time-aware summaries take into account the frequency of terms over time in different versions of web pages to determine salience.

In~\cite{varadarajan2006sqs} the authors create a graph, where the nodes represent text chunks and edges indicate relation between the chunks. However, in contrast to our work, the authors in~\cite{varadarajan2006sqs} consider as optimal summary the maximum spanning tree of the document graph that contains all the keywords; the graph in~\cite{varadarajan2006sqs}  is \emph{not} a character n-gram graph, there is no proposed methodology for the chunking (other than a given parsing delimiter parameter), not all the edges are kept (only those above a given threshold parameter). Furthermore, in~\cite{varadarajan2006sqs} the `TF-IDF'-related Okapi function is used to assign weights to nodes, indicating self-importance of a node. 

In multi-document summarization, different iterative ranking algorithms like PageRank~\cite{brin1998als} and HITS~\cite{kleinberg1999ash} over graph representations of texts have been used to determine the salient terms over a set of source texts~\cite{mihalcea2005mds}. Salience has also been determined by the use of graphs, based on the fact that documents can be represented as `small world' topology graphs~\cite{matsuo2001dsw}, where important terms appear highly linked to other terms. Finding the salient terms, one can determine the containing sentences' salience and create the final summary. 

In another approach~\cite{hendrickx2008using}, content units (sentences) are assigned a normalized value (0 to 1) based on a set of graphs representing different aspects of the content unit. These aspects include: query-relevance; cosine similarity of sentences within the same document (termed \textbf{relatedness}); cross-document relatedness, which is considered an aspect of redundancy; redundancy with respect to prior texts; and coreference based on the number of coreferences between different content units. All the above aspects and their corresponding graphs are combined into one model that assigns the final value of salience using an iterative process. The process spreads importance over nodes based on the `probabilistic centrality' method that takes into account the direction of edges to either augment or penalize the salience of nodes, based on their neighbors' salience.

The notion of Bayesian expected risk (or loss) is applied in the summarization domain by~\cite{kumar2009estimating}, where the selection of sentences is viewed as a decision process, where the selection of each sentence is considered a decision and the system has to select the sentences that minimize the risk.

The CLASSY system (\eg{} see ~\cite{conroy2007cd}, ~\cite{conroy2009csa}) extracts frequently occurring (`signature') terms from source texts, as well as terms from the user query. Using these terms, the system estimates an `oracle score' for sentences, which relates the terms contained within the candidate sentences to an estimated `ideal' distribution based on term appearance in the query, the signature terms and the topic document cluster. Some optimization method (mostly Integer Linear Programming) is then used to determine the best set of sentences for a given length of summary, given sentence weights based on their `oracle score'. 

This article proposes a summarization method that uses language-independent and generic operators that apply to a generic representation of chunks based on interconnected n-grams. 
The `n-gram graph'-based approach has much in common with~\cite{ggianna2008sse}, for chunk (and, thus, sentence) similarity. 
This method overcomes the need for any kind of preprocessing and offers a basic (i.e. core) method for extractive summarization. 
Even the chunking process, which separates a sentence into sub-sentential strings, is based upon statistical analysis of a given document set. The method does not use the bag-of-words approach, as the n-gram graphs take into account the relevant position of n-grams in the text. We do not use any features like sentence position or part-of-speech information. Our method does not deduce salience based on the centrality or connectedness of graph vertices. 

The method extracts a graph expected to represent the common content of input texts, which is in turns considered to indicate salient information. Given a user query, the approach combines --- using n-gram graph operators --- the common content graph with the given query into an overall importance-indicative graph. Then, we calculate the salience of source text sentences based on the similarity of their respective n-gram graph representation to the overall graph, \ie{} the more a sentence representation is similar to the representation of content and query, the more it is considered appropriate for the final summary. Our methodology is described in depth in section \ref{sec:Methodology}.


\subsection{Redundancy and Novelty}
A problem that is somewhat complementary to salience selection is that of \textbf{redundancy detection}. Whereas salience, which is a desired attribute for the information chunks in the summary, can be detected via measuring the similarity of these chunks to a query, redundancy indicates the \emph{unwanted} repetition of information. Research on redundancy has given birth to the Marginal Relevance measure~\cite{carbonell1998umd} and the Maximal Marginal Relevance (MMR) selection criterion. The basic idea behind MMR is that `good' summary sentences (or documents) are sentences (or documents) that are relevant to a topic without repeating information already in the summary. 
The MMR measure is a generic linear combination of any two principal functions that can measure relevance and redundancy. 

Another approach to the redundancy problem is that of the Cross-Sentence Informational Subsumption (CSIS)~\cite{radev2000cbs}, where one judges whether the information offered by a sentence is contained in another sentence already in the summary. The `informationally subsumed' sentence can then be omitted from the summary. The main difference between the two approaches is the fact that CSIS is a binary decision on information subsumption, whereas the MMR criterion offers a graded indication of utility and non-redundancy. 

Other approaches, overviewed in~\cite{allan2003ran}, use statistical characteristics of the judged sentences with respect to sentences already included in the summary to avoid repetition. Such methods are the NewWord and Cosine Distance methods~\cite{larkey2003utc} that use variations of the bag-of-words based vector model to detect similarity between all pairs of candidate and summary sentences. Other, language model-based methods create a language model of the summary sentences, either as a whole or independently, and compare the language model of the candidate sentence to the summary sentences model~\cite{zhang2002nar}. The candidate sentence model with the minimum KL-divergence from the summary sentences' language model is supposed to be the most redundant. 

The CLASSY system~\cite{conroy2007cd,conroy2009csa} represents documents in a term vector space and enforces redundancy through the following process: Given a pre-existing set of sentences $A$ corresponding to a sentence-term matrix $M_A$, and a currently judged set of sentences $B$ corresponding to a matrix $M_B$, $B$ is judged using the term sub-space that is orthogonal to the eigenvalues of the space defined by $A$; this means that only terms that are not already considered important in $A$ will be taken into account as valuable content.

In this work, we have used and evaluated two different strategies for the detection of information redundancy. These strategies use a statistical, graph-based model of sentences by exploiting character n-grams. The first strategy, similarly to CSIS, compares all the candidate sentences and determines the redundant ones. The second strategy, aiming to detect intra-summary novelty and more similar to MMR, creates an iterative n-gram graph model for every snapshot of the summary after a new sentence is added to it. Then, each new candidate sentence is compared to that graph model to determine redundancy. 



\subsection{Graph-based Methods and Graph Matching}
Graphs have been used to determine salient parts of text~\cite{mihalcea2004gbr,erkan2004lgb,erkan2004umd} or query related sentences~\cite{otterbacher2005urw} in close relation to the summarization process. Lexical relationships~\cite{mohamed:qbs2006} or rhetorical structure~\cite{Marcu2000tta} and even non-apparent information~\cite{lamkhede2005msu} have been represented by graphs. Graphs have also been used to detect differences and similarities between source texts~\cite{Mani1997Mds} and inter-document relations~\cite{witte:cbm}, as well as relations of various granularity from cross-word to cross-document as described in Cross-Document Structure Theory~\cite{radev2000cti}. 

In this work, the graphs are used to represent strings of any length or granularity, from chunk to sentence, to document set. Throughout the proposed methodologies, we use a set of different operators, like similarity, merging, intersection, to perform different subtasks of the summarization process (query expansion, content selection, redundancy control). 

Graph similarity calculation methods can be classified in two main categories, according to the literature.
\begin{description}
 \item[Isomorphism-based] Isomorphism is a bijective mapping between the vertex set of two graphs $V_1, V_2$, such that all mapped vertices are equivalent, and every pair of vertices from $V_1$ shares the same state of neighborhood, as their corresponding vertices of $V_2$. In other words, in two isomorphic graphs all the nodes of one graph have their unique equivalent in the other graph, and the graphs have identical connections between equivalent nodes. Based on the isomorphism, a \textbf{common subgraph} between $V_1, V_2$, can be defined as a subgraph of $V_1$ having an isomorphic equivalent graph $V_3$, which is a subgraph of $V_2$. The \textbf{maximum common subgraph} of $V_1$ and $V_2$ is defined as the common subgraph with the maximum number of vertices. For more formal definitions and an excellent introduction to the error-tolerant graph matching, \ie{} fuzzy graph matching, see~\cite{bunke1998etg}.

Given the definition of the maximum common subgraph, a series of distance measures have been defined using various methods for the calculation of the maximum common subgraph, or similar constructs like the Maximum Common Edge Subgraph, or Maximum Common Induced Graph (also see~\cite{raymond2002rcg}).

 \item[Edit-distance Based] Edit distance has been used in fuzzy string matching for some time now, using many variations (see~\cite{navarro2001gta} for a survey on approximate string matching). The edit distance between two strings corresponds to the minimum number of edit character operations (namely insertion, deletion and replacement) needed to transform one string to the other. Based on this concept, a similar distance can be used for graphs~\cite{bunke1998etg}. Different edit operations can be given different weights, to indicate that some edit operations indicate more important changes than others. The edit operations for graphs' nodes are \emph{node} deletion, insertion and substitution. The same three operations can by applied on edges, giving \emph{edge} deletion, insertion and substitution.
\end{description}

Using a transformation from text to graph, the aforementioned graph matching methods can be used as a means to indicate text similarity. A graph method for text comparison can be found in~\cite{tomita2004csb}, where a text is represented by first determining weights for the text's terms using a TF-IDF calculation and then by creating graph edges based on the term co-occurrences. The method proposed in this article does not require term extraction/identification and the corresponding representation graph is constructed by exploiting the text in a direct manner (i.e. no language-dependent preprocessing is required), without exploiting further background/supportive information such as a corpus for the calculation of TF-IDF or any other weighting factor.

The main difference of our method to existing methods is that we enter the sub-word level, through the use of character n-grams. We aim to perform all required tasks towards the summarization of a set of documents using a uniform representation of sentences, senses, documents and document sets, regardless of the underlying language.  Moreover, we map seemingly different summarization subtasks, such as content selection and query expansion, to a set of basic graph operators that function as generic purpose NLP operators over a common representation.

In order to understand the n-gram graph representation we use, one should take into account the fact that adjacency between different linguistic units within specific contexts seems to be a very important factor of meaning. Contextual information has been widely used and several methodologies have been built upon its value (\eg{}~\cite{burgess1998ecs,yarowsky1995uws}). 

Our methodology, described in detail in section \ref{sec:Methodology}, can be summarized by the following main steps:
\begin{itemize}
 \item Analysis of source documents' content.
 \item Query expansion.
 \item Candidate content grading.
 \item Redundancy removal or novelty detection.
 \item Composition of the summary.
\end{itemize}
In the following paragraphs we present the framework (i.e. the representation and the operators) that is used throughout these steps. Having said that, we need to emphasize on the point that a single framework allows for different operators on the Natural Language Processing (NLP) domain. An early presentation of the concepts and processes described herein can be found in~\cite{ggianna2008tun}.

\section{N-gram Graphs, Operators and Algorithms}\label{sec:NGramGraphRepr}
We now provide the definition of n-gram, given a text (viewed as a character sequence):
\begin{definition}
If $n>0, n\in \IZ $, and $c_i$ is the $i$-th character of an $l$-length character sequence $T^{l}=(c_1,c_2,...,c_l)$ (our text), then\\
a \textbf{character n-gram}\index{n-gram!character} $S^{n}=(s_1,s_2,...,s_n)$ is a subsequence of length $n$  of $T^{l} \iff\text{ for }i\in [1, l-n+1]\colon\text{ and }j\in [1,n]\colon s_j=c_{i+j-1}$. We shall indicate the n-gram spanning from $c_i$ to $c_k, k>i$, as $S_{i,k}$, while n-grams of length $n$ will be indicated as $S^{n}$.
\end{definition}

The meaning of the above formal specification, is that n-gram $S^{n}$ can be found as a substring of length $n$ of the original text, spanning from the $i$-th to the $(i+n-1)$-th character of the original text. The length $n$ of an n-gram is called either the \textbf{length, size or the rank of the n-gram}.

The \textbf{n-gram graph} is a graph $G=\{V^G,E^G,L,W\}$, where $V^G$ is the set of vertices, $E^G$ is the set of edges, $L$ is a one-to-one function assigning a label to each vertex and edge and W is a function assigning a weight to every edge. We consider that $L$ labels edges by concatenating the labels of the corresponding vertices in the direction of the edge, if the edge is directed, or in lexicographic order, if the edge is undirected, adding also a special separator character. Labels in vertices are n-grams, $v^G\in V^G$, and the edges $e^G\in E^G$ (the superscript G will be omitted where easily assumed) connecting the n-grams indicate adjacency of the corresponding vertex n-grams in a specific context within distance \dist{}  (also see~\cite{ggianna2008sse}).

\begin{example}
 Vertex $v_1$ corresponds to n-gram ``abc'' and $v_2$ to ``bcd''. Then, $L(v_1)=$``abc'', $L(v_2)=$``bcd''. If $e_1=(v_1,v_2)$ the edge connecting $v_1$ and $v_2$, then $L(e_1)=$``abc$\rightarrow{}$bcd'', where $\rightarrow{}$ is the special separator character.
\end{example}

The edges within this work are weighted by applying the number of co-occurrences of the vertices' n-grams within the given window in the original documents. For simplicity, when for a vertex $v$, $v\in{}V^G$ we may also write $v\in{}G$; the same notation may be used for an edge $e\in{}E^G$, where we may write $e\in{}G$.

Given two instances of n-gram graph representation $G_1,G_2$, there is a number of operators that can be applied on $G_1, G_2$ to provide the n-gram graph equivalent of union, intersection and other such operators of set theory. In our summarization task, these operators are useful as primary tools for all the subtasks --- \ie{} salience detection, novelty detection, redundancy removal, query expansion. An example of such an operator is the merging operator of $G_1$ and $G_2$ corresponding to the union operator in set theory. This operator adds all edges from both operand graphs to a third one, while making sure no duplicate edges are created. Two edges are considered duplicates of each other, when they share identical vertices. We note that the definition of identity between vertices can be customized; within our applications two vertices are the same if they correspond to the same n-gram. 

The definition of the graph operators is actually non-trivial, because a number of questions arise, such as the handling of weights on common edges after a union operator or the `meaning' and thus handling of the zero-weighted edges after the application of any operator.

All the operators that we shall present operate on edges only, because we consider single nodes to be of little value. We argue that information is contained within the \emph{relation between n-grams} and not in the n-grams themselves. Therefore, our minimal unit of interest is the edge, which is actually a pair of vertices. 

\newcommand{\mCont}{\text{contains}}
\newcommand{\mSim}{\text{sim}}
\newcommand{\mGOne}{\ensuremath{G_1}}
\newcommand{\mGTwo}{\ensuremath{G_2}}
\newcommand{\AllGraphs}{\ensuremath{\overline{\GG{}}}}

Overall, we have defined a number of operators all of which --- with the exception of similarity --- are functions from  $\AllGraphs{}\times{}\AllGraphs{}$ to $\AllGraphs{}$, where \AllGraphs{} is the set of valid n-gram graphs of a specific rank $n$. Thus, operators function upon graphs of a given rank and produce a graph of the same rank. The operators are the following.

\begin{itemize}
\item The similarity function\index{n-gram graph!similarity} $\mSim{}:\AllGraphs{}\times{}\AllGraphs{}\rightarrow{}\RR{}$ which returns a value of similarity between two n-gram graphs. This function is symmetric, in that $\mSim(\mGOne{},\mGTwo)=\mSim(\mGTwo,\mGOne)$. There are many variations of the similarity function within this study, each fitted to a specific task. The common-ground of these variations is that the similarity values are normalized in the $[0,1]$ interval, with higher values indicating higher actual similarity between the graphs. The computation of similarity is described in subsection \ref{I:sec:SimilarityAndContainment}.

 \item The containment function\index{n-gram graph!containment} \emph{$\mCont{}$}, which indicates what part of a given graph $\mGOne$ is contained in a second graph $\mGTwo$. This function is expected to be asymmetric. In other words, should the function indicate that a graph $\mGOne{}$ is contained in another graph $\mGTwo{}$, we know nothing about whether the inverse stands. 

In the implementations of the containment function proposed herein, result values are normalized in the $[0,1]$ interval, with higher similarity values indicating that a bigger part of a given graph $\mGOne{}$ is contained in a second graph $\mGTwo{}$. We consider a graph to be contained within another graph if all its edges appear within the latter. If this is not the case, then any common edge contributes to the overall containment function a percentage inversely proportional to the size of \mGOne{}, so that the function value indicates \emph{what part} of \mGOne{} is contained within \mGTwo{}. The computation of containment is described in subsection \ref{I:sec:SimilarityAndContainment}.

 \item The merging or union operator\index{n-gram graph!merging}\index{n-gram graph!union} $\cup{}$, returns a graph with all the edges, both common and uncommon, of the two operand graphs. In our implementation of this operator we have decided that the union operator will set the weight of a common edge equal to the average of the weights into the corresponding graphs. 

 \item The intersection operator\index{n-gram graph!intersection} $\cap{}$, returns a graph with the common edges of the two operand graphs, with averaged weights.
 The averaging over edge weights is based on the idea that the intersection (and the union) of two graphs should be a graph that includes the edges of both operand graphs, with edge weights as close as possible to both the original graphs. 
The union (merge) and intersection operators are presented in section \ref{sec:UnionAndMerging}.

 \item The delta operator\index{n-gram graph!delta operator}\index{n-gram graph!all-not-in operator} (also called \textbf{all-not-in operator}) $\bigtriangleup{}$ returning the subgraph of a graph $G_1$ that is not common with a graph $G_2$. This operator is non-symmetric, \ie{} $G_1\bigtriangleup{}G_2\neq{}G_2\bigtriangleup{}G_1$, in general.

 \item The inverse intersection\index{n-gram graph!inverse intersection} operator $\bigtriangledown{}$ returning all the edges of two graphs that are not common between the graphs. This operator is symmetric, \ie{} $G_1\bigtriangledown{}G_2=G_2\bigtriangledown{}G_1$.
\end{itemize}


Zero-weighted edges are treated as all other edges, even though zero weight means that the edge does not exist (\ie{} the vertices are not related).
The empty graph $\emptyset{}$ is a graph with no nodes and no edges. The size of any graph is its edge count and is indicated as $|\mGOne{}|$ for graph \mGOne{}.

The algorithm we use~\cite{ggianna2008sse} to convert a given string to its character n-gram graph representation is quite simple:
\begin{itemize}
 \item Extract all character n-grams of rank $n$ from a given text and create graph vertices, one for every unique n-gram. The vertices are labeled by their corresponding n-gram.
 \item Add edges connecting all n-grams that occur (at least once) within a given distance \dist{} of each other in the string. In this work, the weight of the added edge is the number of co-occurrences of the corresponding vertices\footnote{After the application of operators between graphs, the edge weights, \eg{} of union graphs, may represent average co-occurrences, or other functions of co-occurrences, thus not being integer numbers.}.
\end{itemize}

\subsection{Similarity and Containment}\label{I:sec:SimilarityAndContainment}
To represent a character sequence or text\index{n-gram graph!text representation} we can use a \emph{set} of n-gram graphs, for various n-gram ranks \index{n-gram!rank}, instead of a single n-gram graph. To compare a sequence of characters in a chunk, a sentence, a paragraph or a whole document (i.e. in any textual chunk), we apply variations of a single algorithm that acts upon the n-gram graph representation of the character sequences. The algorithm is actually a similarity measure between two n-gram graph sets corresponding to two texts $T_1$ and $T_2$. This similarity can be indicative of the similarity of content of two information chunks, in the way any fuzzy string matching technique is. We can therefore apply this similarity measurement to determine whether \eg{} a given sentence is related to a user query of a summarization task.

We consider that an n-gram graph maps to a given n-gram rank in this work, \ie{} the rank $r$ is a parameter of the n-gram graph. Given that the representation of a text $T_i$ is a set of graphs $\mathbb{G}_i$, containing graphs of various ranks, and two texts $T_1,T_2$ we use the \textbf{Value Similarity} $\text{VS}^r$ to compare graphs of the same rank. So, for every n-gram rank $r$ of $\mathbb{G}_1, \mathbb{G}_2$ we use the corresponding graph of rank $r$, \eg{} $\mGOne\in\mathbb{G}_1$ and $\mGTwo\in\mathbb{G}_2$. The $\text{VS}^r(G_1,G_2)$ measures how many of the edges contained in graph $\mGOne$ are contained in graph $\mGTwo$, considering also the weights of the matching edges. In this measure each matching edge $e$ having weight $w_e^1$ in graph $\mGOne$ contributes $\frac{\text{VR}(e)}{\text{max}(|\mGOne|,|\mGTwo|)}$ to the sum, while not matching edges do not contribute (consider that for an edge $e\notin \mGOne$ we define $w_e^1=0$). The \textbf{ValueRatio ($\text{VR}$)} scaling factor is defined as:
\begin{equation}
\label{I:eqVR}
\text{VR}(e) = \frac{\text{min}(w_e^1, w_e^2)}{\text{max}(w_e^1, w_e^2)}
\end{equation}
The equation indicates that the \emph{ValueRatio} takes values in $[0,1]$, and is symmetric. Thus, the full equation for $\text{VS}_r$ is:
\begin{equation}
\label{I:VS}
	\text{VS}(G^{i} ,G^{j})=\frac{\sum_{e\in G^{i}}\frac{\text{min}(w_e^{i}, w_e^{j})}{\text{max}(w_e^{i}, w_e^{j})}}{\text{max}(|G^{i}|,|G^{j}|)}
\end{equation}
$\text{VS}^r$ is a measure converging to 1 for graphs that share edges with similar weights, which means that a value of $\text{VS}^r=1$ indicates perfect match between the compared graphs. Another important measure is the \textbf{Normalized Value Similarity (NVS)}, which is computed as:
\begin{equation}\label{NVS}
 \text{NVS}^r(\mGOne ,\mGTwo)=\frac{\text{VS}^r}{\frac{\text{min}(|\mGOne|,|\mGTwo|)}{\text{max}(|\mGOne|,|\mGTwo|)}} 
\end{equation}
The fraction $\text{SS}^r(\mGOne ,\mGTwo)=\frac{\text{min}(|\mGOne|,|\mGTwo|)}{\text{max}(|\mGOne|,|\mGTwo|)}$, is also called Size Similarity. The NVS is a measure of similarity where the ratio of sizes of the two compared graphs does not play a role.

The overall similarity $\text{VS}^O$of the sets $\GG{}_1, \GG{}_2$ is computed as the weighted sum of the VS over all ranks:
\begin{equation}
\label{overall}
\text{VS}^O(\GG{}_1,\GG{}_2)=\frac{\sum_{r\in [\text{\m{}},\text{\M{}}]}r\times \text{VS}^r}{\sum_{r\in [\text{\m{}},\text{\M{}}]}r}
\end{equation} where $\text{VS}^r$ is the $\text{VS}$ measure for extracted graphs of rank $r$ in $\GG{}$, and \m{}, \M{} are arbitrary chosen minimum and maximum n-gram ranks.

The function \emph{\mCont{}()}\index{n-gram graph!containment} realizing the graph containment operator has a small, but significant difference from the value similarity function: It is not commutative. More precisely, if we call Value Containment ($\text{VC}^r$) the containment function using edge weights, then $\text{VC}^r$ is:
\begin{equation}
\label{I:VC}
\text{VC}^r(\mGOne ,\mGTwo)=\frac{\sum_{e\in \mGOne}\frac{\text{min}(w_e^1, w_e^2)}{\text{max}(w_e^1, w_e^2)}}{|\mGOne|}
\end{equation}

The denominator is the cause for the asymmetric nature of the function and makes it correspond to a graded membership function between two graphs.

\subsection{Graph Union (or Merging) and Intersection}\label{sec:UnionAndMerging}\index{n-gram graph!union}\index{n-gram graph!merging}
The union, or merge, operator $\cup$ has two important aspects. The first deals with unweighted edges as pairs of labeled vertices $e=(v_1,v_2)$, while the second considers the weights of the edges as well. 

The merge operator defines the basic operator required to perform updates in the graph model. The intersection operator, on the other hand, can be used to determine the common subgraphs of different graphs. This use has a variety of applications, such as common topic detection in a set of documents (see section \ref{sec:ContentSelection}) or the detection of `stopword-effect' edges. The `stopword-effect' edges are edges that are apparent in the graphs of most texts of a given language and have high frequency, much like stopwords. 

The detection of \textbf{stopword-effect edges} in n-gram graphs can be accomplished by simply applying the intersection operator upon graphs of (adequately big) texts of various different topics. The resulting graph will represent that part of language that has appeared in all the text graphs and can be considered `noise'. More on the notion of noise in relation to n-gram graphs can be found in section \ref{sec:GraphNoise}.

When performing the union of two graphs we create a graph $\mGOne{}\cup\mGTwo{}=G^u=(V^u,E^u,L,W^u)$, such that $E^u=E^\mGOne{}\cup{}E^\mGTwo{}$, where $E^\mGOne{},E^\mGTwo{}$ are the edge sets of $\mGOne{},\mGTwo{}$ correspondingly. In out implementation we consider two edges to be equal $e_1=e_2$ when they have the same label, \ie{} $L(e_1)=L(e_2)$\footnote{We consider the labeling function to be the same over all graphs.}, which means that the weight is not taken into account when calculating $E^u$.

\index{n-gram graph!union!edge weighting}
To calculate the weights in $G^u$ there can be various functions depending on the effect the merge should have over weights of common edges. One can follow the fuzzy logic paradigm and keep the maximum of the weights of a given edge in two source graphs $W^u(e)=\text{max}(W_1(e),W_{2}(e))$, where $W_{1},W_{2}$ are the weighting functions of the corresponding graphs and $e$ is a common edge of \mGOne{} and \mGTwo{}. Another alternative would be to keep the average of the values so that the weight represents the expected value of the weights of the original weights. Given these basic alternatives, we chose the average value as the default union operator effect on edge weights. It should be noted that the merging operator is a specific case of the graph update function presented in section \ref{I:secDocSetsUpdt}. Formally, if $E^1, E^2$ are the edge sets of $\mGOne,\mGTwo$ correspondingly, $W^u$ is the result graph edge weighting function and $W_1,W_2$ are the weighting functions of the operand graphs with $e\notin{}E^u\Rightarrow{}W^u(e)=0, i\in{1,2}$, then the edge set $E^u$ of the merged graph is:
\begin{equation}
 E^u{}=E^1\cup{}E^2, W^u(e)=\frac{W_1(e) + W_2(e)}{2}, e\in{}(E^1\cup{}E^2)
\end{equation}

The intersection operator\index{n-gram graph!intersection} $\cap$ returns the common edges between two graphs $\mGOne,\mGTwo$ performing the same averaging operator upon the edges' weights. Formally the edge set $E^i$ of the intersected graph is:
\begin{equation}
 E^i{}=\{e|e\in\mGOne\wedge{}e\in\mGTwo\}, W^i(e)=\frac{W_1(e) + W_2(e)}{2}, e\in{}(E^1\cap{}E^2)
\end{equation}

\subsection{Delta (All-not-in) and Inverse Intersection}
The Delta or all-not-in operator\index{n-gram graph!delta operator}\index{n-gram graph!all-not-in operator} $\bigtriangleup{}$ is a non-commutative operator, that given two graphs $\mGOne,\mGTwo$ returns the subset of edges from the first graph, that do not exist in the second graph. Formally the edge set $E^\bigtriangleup{}$ is:
\begin{equation}
 E^\bigtriangleup{}=\{e|e\in\mGOne\wedge{}e\notin\mGTwo\}
\end{equation}
The weight of the remaining edges is not changed when applying the delta operator. Obviously, the operator is non-commutative.

A similar operator is the inverse intersection operator\index{n-gram graph!inverse intersection} $\bigtriangledown{}$ which creates a graph that only contains the \emph{non-common} edges of two graphs. The difference between this operator and the delta operator is that in the inverse intersection the edges of the resulting graph may belong to any of the original graphs. Formally the resulting edge set $E^\bigtriangledown{}$ is:
\begin{equation}
 E^\bigtriangledown{}=\{e|e\in(\mGOne\cup{}\mGTwo)\wedge{}e\notin(\mGOne\cap{}\mGTwo)\}
\end{equation}

Both the delta and the inverse intersection operators can be applied to determine the differences between graphs. This way one can \eg{} remove a graph representing `noisy' content from another graph. Another application is determining the non-common part of two texts that deal with the same subject, which may refer to the unique or novel part of each text, with respect to the subject.

\subsection{Representing Document Sets - Updatability}\label{I:secDocSetsUpdt}
The n-gram graph representation specification indicates how to map a text to an n-gram graph. However, in our task it is required to represent a whole document set. The most simplistic way to do this using the n-gram graph representation would be to concatenate the documents of the set into a single overall document, but this kind of approach would not offer an updatable model --- \ie{} a model that could easily change when a new document enters the set.

In our applications we have used an \textbf{update function} $U$ that is similar to the merging operator, with the exception that the weights of the resulting graph's edges are calculated in a different way. The update function $U(\mGOne,\mGTwo,l)$ takes as input two graphs, one that is considered to be the pre-existing graph $\mGOne$ (i.e. a graph that may have resulted by a sequence of applications of the update operator on an initial graph) and one that is considered to be the new graph $\mGTwo$. The function also has a parameter called the \textbf{learning factor}\index{n-gram graph!update function!learning factor} $l\in[0,1]$, which determines the sensitivity of $\mGOne$ to the changes $\mGTwo$ brings. 

Focusing on the weighting function of the graph resulting from the application of $U(\mGOne,\mGTwo,l)$, the higher the value of learning factor, the higher the impact of the new graph to the existing graph. More precisely, the definition of the weighting performed in the graph resulting from $U$ is:
\begin{equation}
W^i(e)=W^1(e) + (W^2(e)- W^1(e)) \times l
\end{equation}

According to this formula, the value of $l=0$ indicates that $\mGOne$ will completely ignore the new graph. A value of $l=1$ indicates that the weights  of the edges of $\mGOne$ will be assigned the values of the new graph's edges' weights. A value of $0.5$ gives us the simple merging operator.

The $U$ operator allows using the graph as a representation model for a set of documents. This approach is used in our case to represent the \textbf{common content of source documents}. The training step for the creation of the content representation model comprises the initialization of a \textbf{common content graph} with the representation of the first document and the subsequent update of that initial graph with the graphs of the other documents. Especially, when one wants the common content graph's edges to hold weights averaging the weights of all the individual graphs that have contributed to the common content graph, then the $i$-th new graph that updates the common content graph should use a learning factor of
\begin{equation}
 l=1.0-\frac{i-1}{i}, i>1
\end{equation}
This methodology creates a common content graph that can function as a representative graph for all the source documents, in that we expect it to be as `close' as can be to the individual graphs of the individual documents, in terms of value similarity. 

When the common content graph is created, one can determine whether a new document is similar to the content of the source documents by measuring the similarity of the document graph to the common content graph. This information has been used to determine the salience of information chunks in section \ref{sec:ContentSelection}. We have further used the update operator to determine noisy information, in terms of useless graph edges. This is illustrated in the following paragraphs.

\subsubsection{Determining Noise Using the N-gram Graph Operators}\label{sec:GraphNoise}
When determining the common content graph, one faces the presence of noise within the graph. The noise within the graphs for a classification task would be the set of common subgraphs over all classes of documents, as they do not offer distinctive information. Similarly, in our summarization task, we consider that the part of the common content graph that would appear, no matter the underlying topic of the sources, is noise. In traditional text classification techniques stopword removal is usually used to remove what is considered to be the noisy part of the data. Up to this point we have seen that n-gram graphs do not need such preprocessing. However, based on the task, we can usually identify non-interesting parts of data that hinder the task. This `noise' can be removed via the already proposed n-gram graph algorithms.

In order to see what we should do to remove the noise, we will use the related paradigm of a classification task. For the classification task, we consider to have a set of training documents for a number of classes. We consider noise the part of information in the graphs that does not help determine the class of a document. If we manage to find this information, as represented in the graphs, we will be able to remove the noise from the common content graph.

In the case of a classification task, we create a merged graph --- using the update operator --- for the full set of training documents $T_c$ belonging to each class $c$. After creating the classes' \textbf{model graphs}, one can determine the maximum common subgraph between classes and remove it to improve the distinction between different classes. A number of questions arise given this train of thought:
\begin{itemize}
 \item \emph{How can one determine the maximum common subgraph between classes?} According to our operators, it is easy to see that the maximum common subgraph is the intersection of all the class graphs. In other words, the same operator that is used to determine the common content \emph{within} a class of documents, becomes useful as noise indicator when doing \emph{inter-class} analysis.
 \item \emph{Is this (sub)graph unique?} No, it is not. Even though the intersection operator is associative if edge weights are omitted, the averaging of edge weights per intersection causes non-associativity. In other words, $(G_1\cap{}G_2)\cap{}G_3\neq{}G_1\cap{}(G_2\cap{}G_3)$, due to the calculations of the edge weights .
 \item \emph{Can we approximate the noisy subgraph, without iterating through all the classes?} Yes. As can be seen in Figure \ref{NoiseGraphConvergence} the noisy subgraph can be easily approximated in very few steps. In the figure the horizontal axis indicates the number of consecutive intersections performed between the classes' graphs and the vertical axis illustrates the number of edges of the resulting intersected graph. It shows that even from the third iteration there is only insignificant change in the resulting graph size.
\begin{figure}
 \centering
 \includegraphics[width=0.6\textwidth]{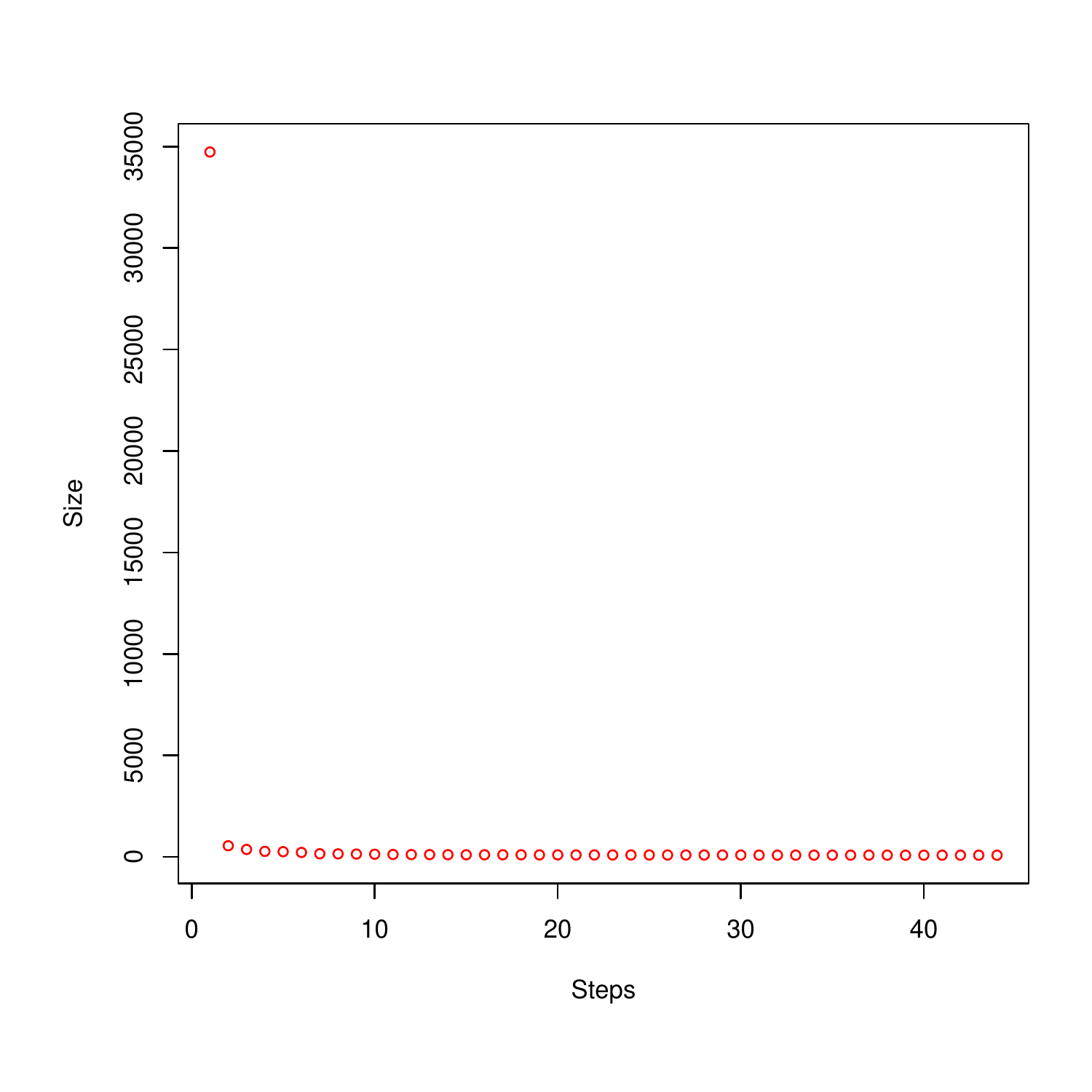}
 \caption{Convergence of common graph size over the number of intersections}\label{NoiseGraphConvergence}
\end{figure}
 \item \emph{Does the removal of the noisy (sub)graph from each class graph really improve results?} The answer is yes, as we will immediately show. 
\end{itemize}

To support our intuition that there is a common part of the n-gram graphs whose effect is the same as that of noise (\ie{} what we have called \textbf{stopword-effect edges}), we performed a set of experiments on classification, which can be easily related to performing topic detection over a set of topics, or to performing common content extraction in the analysis step of our summarization methodology. We created a topic graph for each of the 48 TAC 2008\footnote{See the TAC site, at \url{http://www.nist.gov/tac}, for more.} topics, based on the documents contained within each topic. Then, we tried to classify each of these documents to a corresponding topic. 

\begin{figure}
 \centering
 \subfloat[\emph{Including} noise]{\includegraphics[width=0.5\textwidth]{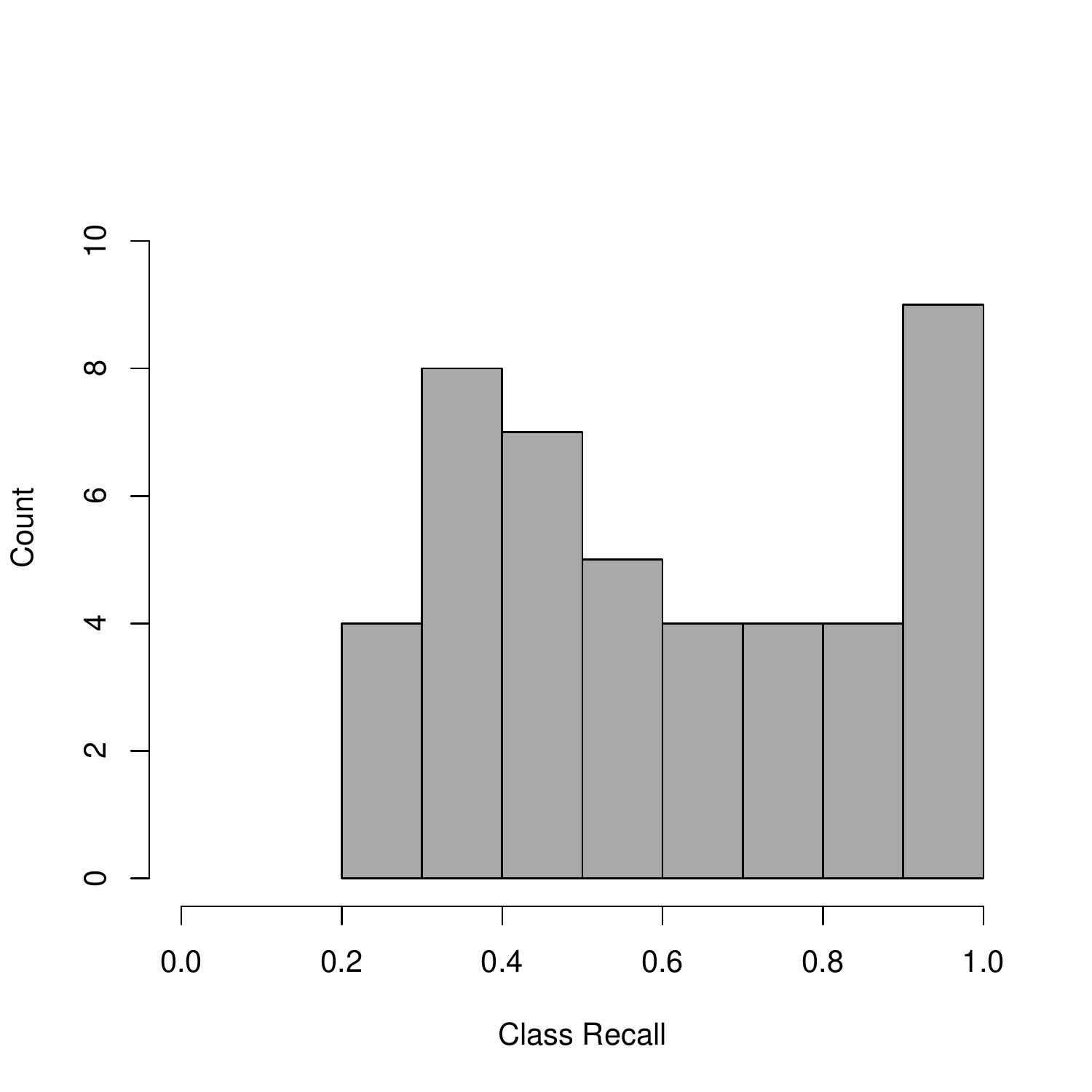}\label{RecallWithNoise}}
 \subfloat[\emph{Without} noise]{\includegraphics[width=0.5\textwidth]{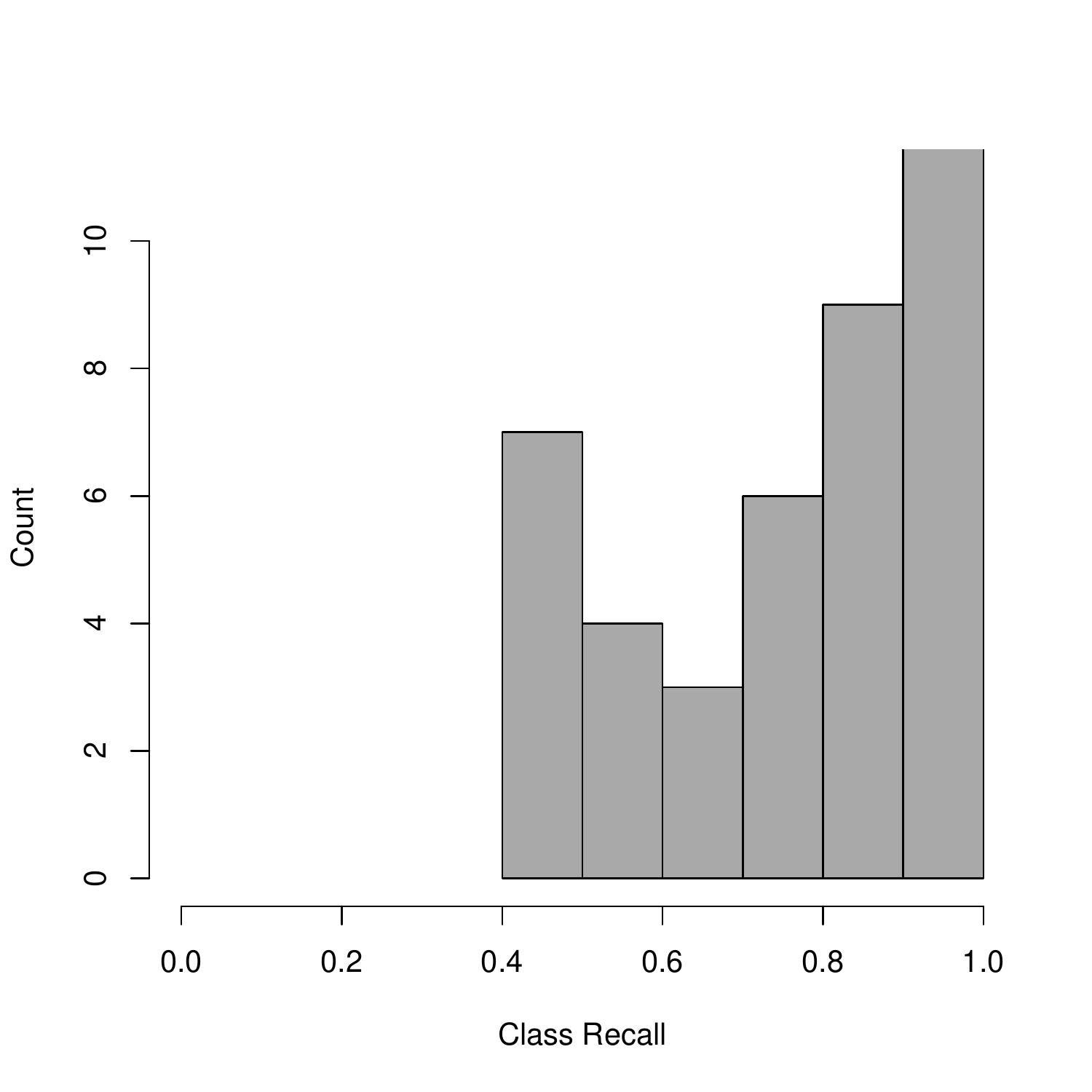}\label{RecallWithoutNoise}}
 \caption{Recall values histogram for all the classes used in the classification task \emph{with} and \emph{without} noise.}
\end{figure}

The classification was performed by measuring the similarity of a judged document to a set of topic-representative graphs. The topic of the graph with the maximum similarity was selected as the topic of the document. This way, we wanted to see if all documents would be found to be maximally similar to their topic graphs, since the training instances would be expected to be recognized as belonging to their original topic. If that was not the case, then perhaps this would be the effect of common elements between the content of different topics, \ie{} noise. As an indication of the performance of the classification we use the recall value, \ie{} the value of the number of documents that were correctly indicated as belonging to a given class, divided by the numbers of all documents that belong to the class. The histogram of recall values for all classes (topics) before removing the noise is shown in figure \ref{RecallWithNoise}. The recall histogram after removing the alleged noise can be seen in figure \ref{RecallWithoutNoise}. The y-axis indicates the number of classes for which the recall value was within a given range (x-axis). Thus, the x-axis indicates the recall value ranges. Of course, ideally, all the values should be $1.0$, to indicate perfect classification. From the figure it is shown that several class results are moved towards higher recall values.

To see whether the results illustrate statistically significant improvement, we used a paired Wilcoxon ranked sum test~\cite{wilcoxon1945icr}, because of the non-normal distribution of differences between the recall of each class in the noisy and noise-free tests. The test indicated that within the 99\% statistical confidence level (p-value $< 0.001$) the results when removing the noise were indeed improved. The conclusion of this experiment is that the removal of noise can really help determine relation to a topic using the n-gram graph representation. This advances the effectiveness of the content selection process when a noise-free topic graph is available.
Therefore, in our content analysis of the original graphs, we make use of the noise removal process to keep a noise-free graph. 

\section{The Proposed Methodology and System: \SYSTEM}\label{sec:Methodology}
\begin{figure}
 \includegraphics[width=0.6\textwidth]{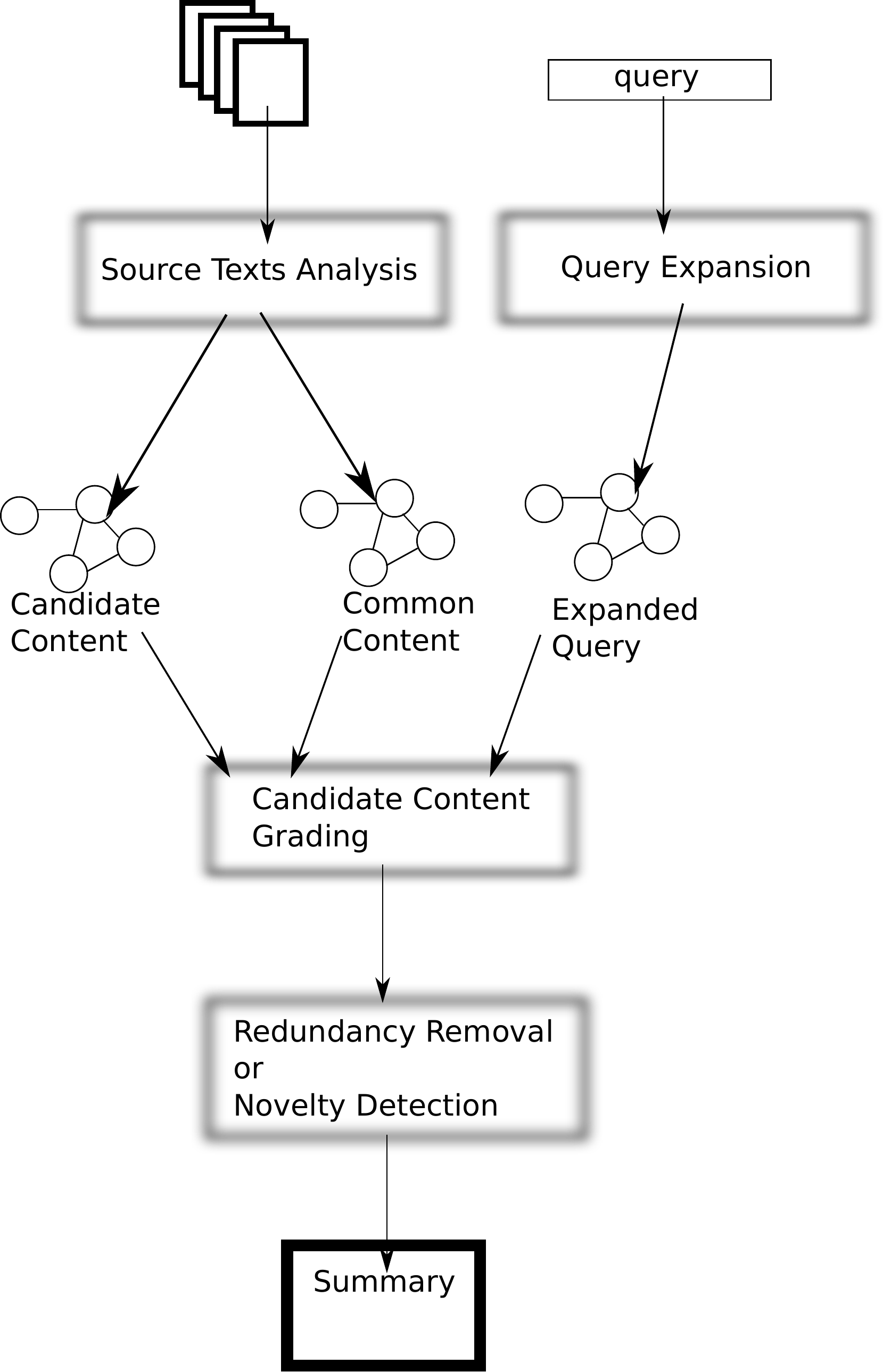}
 \caption{\SYSTEM{} System Overview\label{fig:SystemOverview}} 
\end{figure}

In this section we provide an overview of the \SYSTEM{} system (also see Figure \ref{fig:SystemOverview}), as well as an in-depth analysis of the proposed individual steps.

The \textbf{analysis of source documents' content} step gets as input a set of source documents from which it extracts and represents, through the n-gram graph representation of documents and operators such as the intersection, the information that is common. This information is considered to be important, due to its presence in all the source documents. The output of the step is the common content graph (see more in section \ref{sec:AnalysisOfContent}), representing the common part of the source documents. 

The \textbf{query expansion} step, which is an optional step, aims to annotate a user-supplied query sentence, indicating the subject of the requested summary, with a set of concepts so as to expand it. The input of this step is the query, which is expected to be free-text, and the output is a graph representing the expanded version of the query.

The \textbf{candidate content grading} step assigns scores to sentences from the source documents, in order to evaluate the salience of each sentence. The grading takes into account the user query as well as the information common to all source documents and outputs an ordered list of sentences.

The next step uses either a \textbf{redundancy removal} or a \textbf{novelty detection} approach to avoid repetition in the output summary. This step either eliminates or reranks the already graded and sorted candidate summary sentences, in order to avoid or penalize repetition of information. This step is in fact integrated in the summary creation (final) step. Its input is the set of candidate sentences and its output is a (possibly reranked) subset of the candidate sentences.

Given the output of the previous steps, the final step creates the summary as a sequence of selected candidate sentences. We have not used any sentence ordering method to improve the output of the system, as our main purpose was to determine whether the n-gram graph based tools we devised can be useful throughout the summarization process. This system aims to provide core methods exploiting the n-gram graph representation, providing the basis for more advanced summarization systems. Experiments indicate that indeed, even without the sentence ordering or any rewrite methodology, our system provides promising results.

\subsection{Analysis of Source Documents' Content}\label{sec:AnalysisOfContent}
To analyze the source documents we need to be able to identify and represent the minimal units of information. In other approaches this minimal unit of information would be the word, but we need to remain language independent and also take into account the fact that a word can also be split in sub-word parts. To do this we use the next-character entropy chunking (section \ref{sec:Chunking}). Each chunk, which may be part of a word or longer than a word (\eg{} part of a collocation) is then represented by its corresponding n-gram graph. To identify whether this splitting of the sentence into smaller parts makes any difference in judging the salience of a sentence, we perform experiments (see section \ref{sec:Experiments}) taking into account two alternatives: creating a graph per sentence versus creating a graph per chunk. In the following subsection, we describe the solutions we have applied to identify these chunks.

%

\subsubsection{Next-character Entropy Chunking}\label{sec:Chunking}
To determine the information chunks that will be used in the steps leading to the final summary, we need to determine the appropriate delimiters. To do this in a language-neutral way, we exploit our document corpus to determine the probability $P(c|S_n)$ that a single given character $c$ will follow a given character n-gram $S_n$, for every character $c$ in the corpus. 
The probabilities can then be used to calculate the entropy of the next character for a given character n-gram $S_n$, as follows. If $c_i, i\in\NN{}, 0<i\leq{}M$ the set of characters that have been found to follow the given n-gram $S_n$ and $f_i$ the frequency (count) of $c_i$ being found after $S_n$, then 
\begin{equation}
P_i\equiv{}P(c_i|S_n)=\frac{f_i}{\displaystyle \sum_{i=1}^{M}{f_i}}
\end{equation}

The next-character entropy of $H(S_n)$ in the document corpus is given by:
\begin{equation}
\displaystyle H(S_n)=-\sum_{i=1}^M{P_i\log_2{P_i}}
\end{equation}

\begin{figure}
 \centering
 \includegraphics[width=0.48\textwidth]{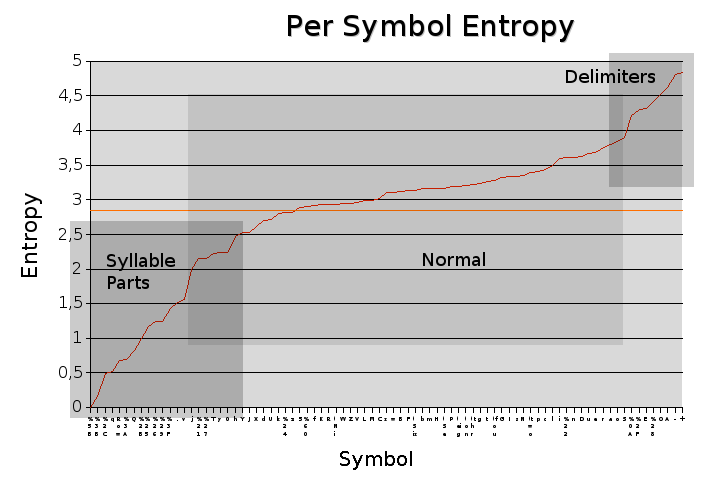}\includegraphics[width=0.48\textwidth]{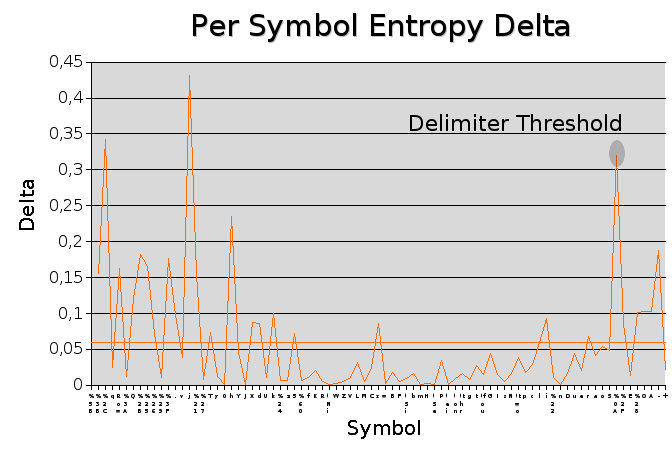}
 \caption{Entropy per Symbol (left) and Delta of Entropy per Symbol (right) - Ordered Descending\label{fig:EntropyPerSymbol}}
\end{figure}

The entropy measure indicates uncertainty. We have supposed that substrings of a character sequence where the entropy of $P(c|S_n)$ surpassed a statistically computed threshold represent candidate delimiters. The threshold is based on the analysis of entropy values that are illustrated in Figure \ref{fig:EntropyPerSymbol} on the left. We noted that one can detect three fuzzy regions in the entropy graph: The first is the region containing delimiters, the second is the region containing non-delimiters and the third contains symbols that have very low entropy of next character, \ie{} they are part of common syllables. The regions are defined by non-trivial changes in the curve of the entropy measure.

To detect the most prominent changes we measured the \textbf{delta} of the entropy values for successive symbols in the graph. The delta $D_H$ is the absolute value of the change in entropy between two successive symbols and is illustrated in Figure \ref{fig:EntropyPerSymbol} on the right. In both figures the horizontal line parallel to the Symbol axes indicates the mean value, of entropy and entropy delta in each figure correspondingly. Given the ordered set $\HH=(H_1, ..., H_M)$ of values $H(S^i_n)$ for $S^1_n, S^2_n, ...$, which the set of n-grams in a corpus, then the delta value $D_H(S^k_n)$ of an n-gram $S^k_n$ with a value $H_k$ as:
\begin{equation}
 \begin{split}
 D_H(H_k)=& |H_{k+1}-H(k)|, k<M\\
 D_H(H_M)=&0
 \end{split}
\end{equation}
where $|x|$ is the absolute value operator.

The entropy value corresponding to the local maximum of the entropy delta in the right half of the symbol-entropy delta function is selected as the threshold (depicted by a dark circle in the figure) for determining delimiter characters. This happens, because we consider exactly three areas, as is depicted in the left part of figure \ref{fig:EntropyPerSymbol}, and we expect their split points to be on either side of the middle symbol in the delta-ordered list of symbols. Formally, the threshold $D_{H,0}$ is given by:

\begin{equation}
 \displaystyle D_{H,0}=\argmax_{i>\lfloor{}\frac{M}{2}\rfloor{}}{H_i}
\end{equation}
where $\lfloor{}x\rfloor{}$ is the floor operator, returning the closest integer lower than or equal to $x$.

In our application we have only checked for unigrams, \ie{} simple letters, as delimiters, even though delimiters of higher rank can be determined. For example, in bi-gram chunking the sequence `,\_' (comma and space) would be considered to be delimiter, while in unigrams the space character only would be considered delimiter. Given a character sequence $S_n$ and a set of delimiters $\DD{}$, our chunking algorithm splits the string after every occurrence of a delimiter $d\in{}\DD$. This way a sentence is split into a set of chunks, that can be then assigned salience, during the content selection process.

\subsection{Query Expansion}\label{sec:QE}
Summarization systems like the ones presented within the DUC and TAC communities need to be able to respond to specific queries. In information retrieval, for some time now, the use of query expansion has been shown to be useful at times. We wanted to try using query expansion in this summarization system as well to determine whether it offers improvement to the system performance.

To perform query expansion we use a two step process. First, we devise a methodology that can map a sentence to a set of concepts, provided external knowledge. Once again, our analysis of the sentence tries to remain as language-independent as possible, even though the use of the external resource may dictate a specific language. The second step for the query expansion is the use of the concepts descriptions mapped to the query sentence, in order to expand the graph representation of the query. We elaborate on these two steps in the following sections.

\subsubsection{Mapping a Sentence to a Set of Concepts Using External Knowledge}\label{III:sec:SentenceToConcepts}
Within the scope of our work we tried to map sentences to concepts (\ie{} to terms with well defined meaning). Usually this happens by looking up words in thesauri (\eg{} WordNet~\cite{miller1990iwl}). In our approach we have used a decomposition module based on the notion of the \textbf{symbolic graph}. 

A symbolic graph is a graph where each vertex contains a string and edges are connecting vertices in a way indicative of a \emph{substring} relation. As an example, if we have two strings \emph{abc, ab} labeling two corresponding vertices, then since \emph{ab} is a substring of \emph{abc} there will be a directed edge connecting \emph{ab} to \emph{abc}\footnote{The symbolic graph can also be represented more efficiently as a trie~\cite{fredkin1960trie}.}. In general, the symbolic graph of a given text $T$ contains every string in $T$ and for every string it illustrates the set of substrings that compose it. 

When a symbolic graph has been constructed, then one can run through all the vertices of the graph and look up each vertex in a thesaurus to determine if there is a match with an entry. If the thesaurus contains an entity lexicalized by the vertex string, then the vertex is assigned the corresponding term \emph{meaning}. In cases of polysemous terms, then the vertex is annotated with all possible meanings of this term. 
This annotated graph, together with a facility that supports comparing meanings is what we call the \textbf{semantic index}.

\newcommand{\mSimMean}{\ensuremath{\text{rel}_\text{Meaning}}}
The semantic index, therefore, represents links between n-grams and their semantic counterparts, implemented as \eg{} WordNet definitions which are textual descriptions of the possible senses. Such definitions are used within Example \ref{III:ex:sensecompare}. If $D_1,D_2$ are the sets of definitions of two terms $t_1,t_2$, then to compare the semantics (meaning) $m_1,m_2$ of $t_1,t_2$ using the semantic index, we actually compare the n-gram graph representation $G_{1i},G_{2j}, 1\leq{}i\leq{}|D_1|, 1\leq{}j\leq{}|D_2|$ of each pair of definitions of the given terms. Within this section we consider the meaning of a term to map directly to the set of possible senses the term has. The relatedness of meaning $\mSimMean{}$ is considered to be the \emph{averaged sum of the similarities} over \emph{all pairs of definitions of the compared terms when represented as n-gram graphs}:
\begin{equation}\label{III:eq:simmeaning}
 \mSimMean(t_1,t_2)=\frac{\sum{}_{G_{1i},G_{2j}} {\mSim{}(G_{1i},G_{2j})}} {|D_1|\times{}|D_2|}
\end{equation}

This use of relatedness implies that uncertainty concerning the actual meaning of terms is handled within the measure itself, because many alternative senses, \ie{} high $|D_1|,|D_2|$, will cause a lower result of relatedness. An alternative version of the relatedness measure, that only requires a single pair to be similar to determine high relatedness of the meanings is the following.
\begin{equation}
 \mSimMean{}'(t_1,t_2)=\text{max}_{G_{1i},G_{2j}}\mSim{}(G_{1i},G_{2j})
\end{equation}
Within our examples in this section we have used equation \ref{III:eq:simmeaning}.

\begin{example}\label{III:ex:sensecompare}
\scriptsize
Compare: smart, clever

WordNet sense definitions for `clever':
\begin{itemize} 
 \item cagey, cagy, canny, clever -- (showing self-interest and shrewdness in dealing with others; ``a cagey lawyer''; ``too clever to be sound'')
 \item apt, clever -- (mentally quick and resourceful; ``an apt pupil''; ``you are a clever man...you reason well and your wit is bold''-Bram Stoker)
 \item clever, cunning, ingenious -- (showing inventiveness and skill; ``a clever gadget''; ``the cunning maneuvers leading to his success''; ``an ingenious solution to the problem'')
\end{itemize}

WordNet sense definitions for `smart':
\begin{itemize}
\item smart -- (showing mental alertness and calculation and resourcefulness)
\item chic, smart, voguish -- (elegant and stylish;``chic elegance'';``a smart new dress'';``a suit of voguish cut'')
\item bright, smart -- (characterized by quickness and ease in learning;``some children are brighter in one subject than another'';``smart children talk earlier than the average'')
\item fresh, impertinent, impudent, overbold, smart, saucy, sassy, wise -- (improperly forward or bold;``don't be fresh with me'';``impertinent of a child to lecture a grownup'';``an impudent boy given to insulting strangers'';``Don't get wise with me!'')
\item smart -- (painfully severe;``he gave the dog a smart blow'')
\item smart -- (quick and brisk;``I gave him a smart salute'';``we walked at a smart pace'')
\item smart -- (capable of independent and apparently intelligent action;``smart weapons'')
\end{itemize}

Relatedness of meaning (\mSimMean{}):
0.0794
\end{example}

In table \ref{III:ex:meaningPairs}, we present some more pairs of terms and their corresponding relatedness values. These preliminary results indicate that, even though the measure appears to have higher values for terms with similar meaning, it may be biased when two words have similar spelling. This happens because the words themselves appear in the definitions, causing a partial match between otherwise different definitions.
\begin{table}
\scriptsize
\centering
\begin{tabular}{l|l||r}
$t_1$ & $t_2$ & $\mSimMean$ \\ \hline
smart & pretty & 0.0339 \\
smart & stupid & 0.0388 \\
run & jump & 0.0416 \\
run & walk & 0.0436 \\
run & die & 0.0557 \\
run & operate & 0.0768 \\
hollow & holler & 0.0768 \\
hollow & empty & 0.0893 \\
\end{tabular}
\caption{Other examples of comparisons in ascending relatedness value: using sense overview\label{III:ex:meaningPairs}}
\end{table}

The results further depend heavily on the textual description --- \ie{} definition --- of any term's individual sense (synset in WordNet). The results from the examples when using  synonyms as descriptors of individual senses can be seen in table \ref{III:ex:meaningPairsSynsOnly}. We notice that \eg{} the words `smart' and `clever' are found to have no relation whatsoever, because no common synonyms are found within  WordNet.
Furthermore, since a given word always appears in its own synonym list, word spelling similarity still plays an important role when judging relatedness between senses of two different words. For example, the words `hollow' and `holler' have an important overlap in terms of spelling. This makes some of their senses have high overlap, which then biases the comparison process to consider the corresponding senses related. 

\begin{table}
\scriptsize
\centering
\begin{tabular}{l|l||r}
$t_1$ & $t_2$ & $\mSimMean$ \\ \hline
smart & clever & 0.0000 \\
run & jump & 0.0017 \\
smart & stupid & 0.0020 \\
run & walk & 0.0020 \\
smart & pretty & 0.0036 \\
run & die & 0.0152 \\
hollow & empty & 0.1576 \\
run & operate & 0.2162 \\
hollow & holler & 0.3105 \\
\end{tabular}
\caption{Other examples of comparisons in ascending relatedness value: using only synonyms from overview\label{III:ex:meaningPairsSynsOnly}}
\end{table}

The use of a semantic index is that of a meaning look-up engine. We remind the reader that the semantic index is actually an annotated symbolic graph. If there is no matching vertex in the graph to provide a meaning for a given input string, then the string is considered to have the meaning of its closest, in terms of graph path length, substrings that have been given a meaning. This `inheritance' of meaning from short to longer strings is actually based on the intuition that a text chunk contains the meaning of its individual parts. Furthermore, a word may be broken down to elementary constituents that offer meaning. If one uses an ontology or even a thesaurus including prefixes, suffixes or elementary morphemes and their meanings to annotate the symbolic graph, then the resulting index becomes a powerful semantic annotation engine, in that it can combine the meaning of sub-word items to determine the word meaning. On the other hand, since composite words or collocations are not necessarily the simple composition of the meaning of their parts, in several cases we expect the resulting meaning annotation to be false.

In the context of this work, we have combined a symbol graph and WordNet into a semantic index to annotate queries with meanings and perform query expansion, knowing that the process will be non-optimal. 

\subsubsection{Query Expansion Through Graph Merging}
Query expansion is based on the assumption that a set of words related to an original query can be used as part of the query itself to improve the set of the returned results. In the literature much work has indicated that query expansion should be carefully applied in order to improve results~\cite{voorhees1994qeu,qiu1993cbq}.

In our approach, we have used a semantic index based on \emph{WordNet's `overview of senses'}. An example of such `overview of senses' for the words `test', `ambiguous' can be seen in Example \ref{III:ex:overview}.

\begin{example}\label{III:ex:overview}
\scriptsize
\begin{verbatim}
Overview of verb test

The verb test has 7 senses (first 3 from tagged texts)

1. (32) test, prove, try, try out, examine, essay 
	-- (put to the test, as for its quality, 
	or give experimental use to; 
	"This approach has been tried with good results"; 
	"Test this recipe")
2. (9) screen, test 
	-- (test or examine for the presence of disease or 
	infection; "screen the blood for the HIV virus")
3. (4) quiz, test -- (examine someone's knowledge of something; 
	"The teacher tests us every week"; 
	"We got quizzed on French irregular verbs")
4. test -- (show a certain characteristic when tested; 
	"He tested positive for HIV")
5. test -- (achieve a certain score or rating on a test; 
	"She tested high on the LSAT and was admitted to 
	all the good law schools")
6. test -- (determine the presence or properties of (a substance))
7. test -- (undergo a test; "She doesn't test well")

Overview of adj ambiguous

The adj ambiguous has 3 senses (first 3 from tagged texts)

1. (9) equivocal, ambiguous -- (open to two or more interpretations; 
	or of uncertain nature or significance; 
	or (often) intended to mislead; "an equivocal statement"; 
	"the polling had a complex and equivocal (or ambiguous) message 
	for potential female candidates"; 
	"the officer's equivocal behavior increased the victim's uneasiness"; 
	"popularity is an equivocal crown"; 
	"an equivocal response to an embarrassing question")
2. (4) ambiguous -- (having more than one possible meaning; 
	"ambiguous words"; "frustrated by ambiguous instructions, 
	the parents were unable to assemble the toy")
3. (1) ambiguous -- (having no intrinsic or objective meaning; 
	not organized in conventional patterns; 
	"an ambiguous situation with no frame of reference"; 
	"ambiguous inkblots")
\end{verbatim}
\end{example}


For a given word $w$ in the original query, the overview of senses is returned by the semantic index; from these senses $s_i, i>0$ we only utilize senses $s_j$ with graph representations $G_{s_j}$ whose (graph) similarity to the common content graph $C_\UU{}$ is greater than a given threshold (see section \ref{I:sec:SimilarityAndContainment} for the definitions of the used functions):
\begin{equation}
G_{s_j}\cap{}C_\UU{}\neq{}\emptyset{} \text{ and }  \text{VS}^O(G_{s_j}, C_\UU{}) > t, t\in{}\RR^+.
\end{equation}
We remind the reader that the content graph is the intersection of all the graph representations of the texts in the set. 

Finally, the query is expanded by merging the representation of the original query $G_q$ and the representations $G_{s_j}$ of all the $j$ senses that have been `filtered' according to the procedure, giving a new query-based content definition $C_\UU{}'$. Having calculated $C_\UU{}'$, we can judge the importance of sentences simply by comparing the graph representation of each sentence to the $C_\UU{}'$. We refer to the removal of the `irrelevant' definitions from the overview of senses as \emph{sense filter}.

Even though the query expansion process was finally rather successful, in our first attempt for query expansion noise was added due to chunks like `an', `in' and `o' which were directly assigned the WordNet meanings of `angstrom', `inch' and `oxygen' correspondingly. This lowered the evaluation scores of our submitted runs in TAC 2008~\cite{ggianna2008tun}. Using the sense filter, the deficiency has been tackled in the current version of the system, significantly improving overall performance.

\subsection{The Content Selection Process}\label{sec:ContentSelection}
Concerning the content matching part of the presented summarization system, the following basic assumptions have been made.
\begin{itemize}
 \item The \textbf{content $C_\UU{}$} of a \textbf{text set (corpus) $\UU{}$} is considered to be the noise-free intersection of all the graph representations of the texts in the set: $C_\UU{}=\bigcap{}^{t\in\UU{}}G_t$, where $G_t$ is the graph representation of text $t$, over all the (arbitrary selected) n-gram ranges. In other words, we consider that the common part of the graph representation of all the texts in a topic, indicates the common content of the texts. Through the (optional) query expansion process one can finally use, instead of $C_\UU{}$, the query-based content definition $C_\UU{}'$, which is determined as described in the previous section.
 \item A sentence $S$ is considered more similar to the content $C_\UU{}$ of a text set, as more of the sentence's chunks (sub-strings of a sentence) have an n-gram graph representation similar to the corresponding content representation. Every chunk's similarity to the content is added to the overall similarity of a sentence to the content. The chunks of a sentence are extracted using the aforementioned entropy-based approach (section \ref{sec:Chunking}). As we will see in the experiments in section \ref{sec:Experiments}, we test whether the use of chunks differentiates the performance of the system as opposed to the case that each sentence is considered a single chunk.
\end{itemize}

According to the chunking process, each sentence is assigned a score, which is actually the sum of the Normalized Value Similarities (see Eq. \ref{NVS}) of its chunks to the content. This process offers an \emph{ordered list of sentences} $\LL{}$. A naive selection algorithm would then select the highest-scoring sentences from the list, until the summary word count limit is reached. However, this would not take redundancy into account and, thus, this is where redundancy removal comes in.

\subsection{Redundancy Removal and Novelty Detection}
In the composition of the final summary, the step of redundancy control is fully integrated. As we will shortly describe, two alternatives were studied concerning the control of redundant information: One detects novelty of a new piece of information with respect to the current summary snapshot or to a user model; the other detects redundancy among the whole set of source information pieces, before the creation of the summary. These approaches just represent two different views of the same problem, which is the control of redundant information within a multi-document summary.

The \emph{novelty} detection process has two aspects, the \textbf{intra-summary} novelty and the \textbf{inter-summary} or \textbf{user-modeled} novelty. 

The intra-summary novelty refers to the novelty of a sentence in a summary, given the content of the summary at a specific time point. 
In order to ensure \emph{intra-summary novelty}, one has to make sure that every sentence minimally repeats already existing information. To achieve this goal, we use the following process:
\begin{enumerate}
 \item Extract the n-gram graph representation of the summary so far, indicated as $G_{\text{sum}}$.
 \item Keep the part of the summary representation that does not contain the common content of the corresponding document set $\UU{}$, $G_{\text{sum}}'=G_{\text{sum}}\bigtriangleup{}C_{\UU{}}$.
 \item For every candidate sentence in $\LL{}$ that has not been already used
 \begin{enumerate}
	\item extract its n-gram graph representation, $G_{cs}$.
	\item keep only $G_{cs}'=G_{cs}\bigtriangleup{}C_{\UU{}}$, because we expect to judge redundancy for the part of the n-gram graph that is not contained in the common content $C_\UU{}$.
	\item assign the similarity between $G_{cs}', G_{\text{sum}}'$ as the sentence redundancy score.
 \end{enumerate}
\item For all candidate sentences in $\LL{}$
\begin{enumerate}
 \item Set the score of the sentence to be its rank based on the similarity to $C_\UU{}$ minus the rank based on the redundancy score.
\end{enumerate}
 \item Select the sentence with the highest score as the best option and add it to the summary.
 \item Repeat the process until the word limit has been reached or no other sentences remain.
\end{enumerate}

The inter-summary or user-modeled novelty refers to the novelty of information apparent when the summarization process takes into account information already available to the reader (as per the TAC 2008 update task). 
This information can be contained in a user model, keeping track of the most recent summaries provided to that user. In the TAC 2008 summarization task, systems are supposed to take into account the first of two sets per topic, set A, as prior user knowledge for the summary of set B of the same topic. In fact, set A contains documents concerning a news item (\eg{} Antarctic ice melting) that have been published before the documents in set B. We have used the content of the given set A, $C_\UU{}_A$, in the redundancy removal process considering it to be the pre-existing user model. To do that we always merged the representation of set A to the representation of the current snapshot of the summary. In other words, the content of set A appears to always be included in the current version of the summary and, thus, new sentences avoid redundancy with respect to A. 

Within this work we have also implemented a method of \emph{redundancy} removal, as opposed to novelty detection, where redundancy is pinpointed within the original set of candidate sentences: We consider a sentence to be redundant, if it surpasses an empirically computed threshold --- in the experiments this threshold had a value of 0.2\footnote{The threshold should be computed via experiments or machine learning to relate with human estimated redundancy of information, but this calculation has not been performed in this work.} --- of overlap to any other candidate sentence. In each iteration within the redundancy removal process each sentence is compared only to sentences not already marked as redundant. As a result of this process, only the sentences that are not marked as redundant are used in the output summary.



Given the whole set of tools we described so far, we now provide some experimental results of applying variations of the aforementioned methodology and the conclusions reached.

\section{Experiments}\label{sec:Experiments}
We conducted numerous experiments using the TAC 2008 corpus. We consider each variation of our system based on a different parameter set to be a different system, with a different System ID. Our main target is to see how our components affect the summarization process as a whole and not to judge individual steps separately. Ideally, the summaries should be judged by humans, but there are automatic methods~\cite{lin2004rpa,hovy2005be,ggianna2008sse} that correlate well to human judgment .

We have used the AutoSummENG~\cite{ggianna2008sse} as our system evaluation method, since it consistently correlates well to the DUC and TAC manually-assigned \textbf{responsiveness measure}. Responsiveness, first appeared in the Document Understanding Conference (DUC) of 2005. This extrinsic measure has been used in later DUCs as well. In DUC\index{DUC} 2005, the appointed task was the synthesis of a 250-word, well-organized answer to a complex question, where the data of the answer would originate from multiple documents~\cite{dang2005od}. In DUC 2005, the question the summarizing `peers', \ie{} summarizer systems or humans, were supposed to answer consisted of a topic identifier, a title, a narrative question and a granularity indication, with values ranging from `general' to `specific'. The responsiveness score is an extrinsic measure that was supposed to provide, as Dang states in~\cite{dang2005od}, a `coarse ranking of the summaries for each topic, according to the amount of information in the summary that helps to satisfy the information need expressed in the topic statement, at the level of granularity requested in the user profile'. 

In the `Automatically Evaluating Summaries of Peers ' (AESOP) task of TAC 2009, the AutoSummENG method was shown to still be one of the top-performing methods in terms of correlation to responsiveness~\cite{dang2009otSLIDES,giannakopoulos_n-gram_2009}. Thus, the evaluation using the automatic AutoSummENG measure is meant to result in a partial ordering, indicative of how well a given summary answers a given question, taking into account the specifications of the question. In the following paragraph we define the task upon which \SYSTEM{} was evaluated using AutoSummENG.

In TAC 2008 there were two tasks. The main task was to produce a 100-word summary from a set of 10 documents (Summary A). The update task was to produce a 100-word summary from a set of subsequent 10 documents, with the assumption that the information in the first set is already known to the reader (Summary B)\footnote{See \url{http://www.nist.gov/tac/publications/2008/presentations/TAC2008_UPDATE_overview.pdf} for an overview of the Text Analysis Conference, Summarization Update Task of 2008.}. There were 48 topics with 
20 documents per topic in chronological order. Each summary was to be extracted based on a topic description defined by a title and a narrative query. For every topic 4 model summaries were provided for evaluation purposes.

At this point we indicate the pitfalls in using an overall evaluation measure like AutoSummENG, ROUGE~\cite{lin2004rpa} or Basic Elements~\cite{hovy2005be} (also see~\cite{belz_thats_2009} for a related discussion):
\begin{itemize}
 \item Small variations in system performance are not indicative of real performance change, due to statistical error.
 \item The measure can say little about \emph{individual summaries}, because it correlates really well when judging a \emph{system}.
 \item The measure cannot judge performance of intermediate steps, because it judges the output summary only.
 \item The measure can only judge the summary with respect to the given model summaries. 
\end{itemize}
Given the above restrictions, we have performed experiments to judge the change in performance when using:
\begin{itemize}
 \item chunk salience scoring versus sentence salience scoring.
 \item redundancy removal versus novelty detection.
 \item query expansion versus no query expansion.
\end{itemize}

\begin{table}
 \centering
 \begin{tabular}{l||c|c||c|c||c|c||r}
System ID & CS & SS & RR & ND & QE & NE & Score \\ \hline
1 & & \checkmark & & \checkmark & & \checkmark & 0.1202 \\ 
2 & & \checkmark & \checkmark & & & \checkmark & \textbf{0.1303} \\ 
3 & \checkmark  & & \checkmark & & \checkmark & & 0.1218 \\ 
4 & & \checkmark & & \checkmark & \checkmark & & 0.1198 \\ 
5 & & \checkmark & \checkmark & & \checkmark & & 0.1299 \\ 
6 & \checkmark & & & & & \checkmark & 0.1255 \\ 
 \end{tabular}
 \caption[Summarization Performance for various settings]{\label{III:tbl:ASPDS}AutoSummENG summarization performance for different settings concerning scoring, redundancy and query expansion. \scriptsize{\textbf{Legend} CS: Chunk Scoring, SS: Sentence Scoring, RR: Redundancy Removal, ND: Novelty Detection, QE: Query Expansion, NE: No Expansion. Best performance in \textbf{bold}.}}
\end{table}

In addition to the results of applying the different system configurations on the summarization task, indicated in table \ref{III:tbl:ASPDS}, we performed an ANOVA (analysis of variance) test to determine whether the System ID --- \ie{} system configuration --- is an important factor for the AutoSummENG similarity of the peer text to the model texts. It was shown (with a p-value below $10^{-15}$) that:
\begin{itemize}
 \item There are topics of various difficulty and the topic is an important factor for system performance.
 \item Selection of different components for the summarizer, from the range of our proposed components, can affect the summaries' quality. The finding was in fact that the SystemID is an important factor of the performance.
\end{itemize}

The systems using chunk scoring have no statistically significant difference in performance from the ones that use sentence scoring, as the (paired) t-test gave a p-value of 0.64. However, the systems using chunk scoring, namely systems 3 and 6, had a slightly lower average performance than the others. The systems using redundancy removal appear to have statistically significant difference in performance from the ones that use novelty detection, nearly at the 0.05 confidence level (one-sided t-test). System 6 was chosen to not use any redundancy removal method and performs near the average of all other systems, thus no conclusion can be drawn. Concerning query expansion, it was not proved whether query expansion indeed offers improvement, as the t-test gave a p-value of 0.74. This result is consistent with the `slight improvement' indicated in~\cite{conroy2006back}, where the reverse mapping of a Porter stemmer was used to expand the query with other versions of its words (\eg{} adding `s' as verb suffix or noun suffix to terms in the query). Similar, non-decisive, results were found by~\cite{blake-unc} where query expansion was determined to be of little use, after experiments were applied.


In table \ref{tbl:TAC2008Experiments} information on the average performance of TAC 2008 participants over all topics is illustrated. More on the performance of TAC 2008 systems can be found in~\cite{dang2008ous}. Our system performs slightly below average but quite better than the last successful participant.

This is very encouraging for the potential of the proposed summarization method, as it is based on generic algorithms performed on a generic representation, providing core operators for addressing the difficulties in each single summarization step. Moreover, the language neutrality of the method shows that it may provide a steady basis, which can be made more effective when combined with heuristics and machine learning methods exploiting language-dependent characteristics. 

%
\begin{table}
 \centering
 \begin{tabular}{l|c}
  \emph{System (TAC 2008 SysID)} & \emph{AutoSummENG Score} \\ \hline
  Top Peer (43) & 0.1991 \\ 
  Last Peer (18) & 0.1029 \\ 
  Peer Average (All Peers) & 0.1648 {\small (Std. Dev. $0.0216$)} \\ \hline 
  \textbf{Proposed System (-)} & \textbf{0.1303} 
 \end{tabular}
 \caption{AutoSummENG performance data for TAC 2008. \scriptsize{NOTE: The top and last peers are based on the AutoSummENG measure performance of the systems.}}\label{tbl:TAC2008Experiments}
\end{table}

To further examine the performance of our system in other corpora, we performed summarization using the configuration that performed optimally in the TAC 2008 corpus on the corpora of DUC year 2006. Systems in DUC 2006 were to synthesize from a set of 25 documents a brief, well-organized, fluent answer to a non-trivially expressed declaration of a need for information. This means that the query could not be answered by just stating a name, date, quantity, or similar singleton. The organizers of DUC 2006, NIST, also developed a simple
baseline system that returned all the leading sentences of the `TEXT' field of the most recent document for each topic, up to 250 words~\cite{dang2006od}.

In Table \ref{tbl:DUC2006Experiments} we illustrate the performance of our proposed system on the DUC 2006 corpus. It is shown that the system strongly outperforms the baseline system, and is less than a standard deviation ($0.0170$) below the AutoSummENG mean performance ($0.1842$) of all the 35 participating systems.

\begin{table}
 \centering
 \begin{tabular}{l|c}
  \emph{System (DUC 2006 SysID)} & \emph{AutoSummENG Score} \\ \hline
  Baseline (1) & 0.1437 \\ 
  Top Peer (23) & 0.2050 \\ 
  Last Peer (11) & 0.1260 \\ 
  Peer Average (All Peers) & 0.1842 {\small (Std. Dev. $0.0170$)} \\ \hline 
  \textbf{Proposed System (-)} & \textbf{0.1783} 
 \end{tabular}
 \caption{AutoSummENG performance data for DUC 2006. \scriptsize{NOTE: The top and last peers are based on the AutoSummENG measure performance of the systems.}}\label{tbl:DUC2006Experiments}
\end{table}

From the comparison between the results on the DUC 2006 and the TAC 2008 task we can conclude that our proposed system performed better in terms of responsiveness in the generic summarization task of DUC 2006 than in the update task of TAC 2008. 
To identify the exact defects of the TAC summaries is non-trivial and requires further investigation, across several dimensions:
\begin{itemize}
 \item What are the problems that affect the performance? Is it the content selection, the ordering of sentences, the anaphora problems, the lack of coherence, or something else?
 \item Can the problem or error be quantified? If yes, how? If not, can a qualitative ranking of quality be applied and, then, approximated by a some evaluation methodology?
 \item How can we minimize the error or maximize the quality?
\end{itemize}

Nevertheless, it is very important that the proposed summarization components offered competitive results \emph{without using machine learning techniques combined with a rich set of sentence features}, like sentence position or grammatical properties. This indicates the usefulness of n-gram graphs as well as the generality of application of the n-gram graph operators and functions. However, other components need to be added to reach state-of-the-art performance, given the existing means of evaluation. These components should aim to improve the overall coherence of the text and tackle problems of anaphora resolution (for examples of such problems see the summary in the appendix section \ref{apd:Sample}).

\section{Discussion and future work}\label{sec:Discussion}
We have offered a generic method, based on the language-neutral representation and algorithms of n-gram graphs, aiming to tackle a number of automatic summarization problems:
  \begin{description}
 \item[Salience detection] We have indicated ways to determine the content of a cluster of documents and judge salience for a given sentence.
 \item[Redundancy removal] We have presented two different approaches following the CSIS and MMR paradigms.
 \item[Query expansion] We have proposed a scheme to broaden a given query, with a slightly improving effect over the summaries. The query expansion module is partially dependent on the language, in that it requires a thesaurus in the same language as the original query to perform expansion.
\end{description}

From the alternatives we examined within the experiments, as far as responsiveness of the summaries is concerned, it stands that:
\begin{itemize}
 \item whole-sentence scoring should be preferred to chunk-based sentence scoring.
 \item query expansion does not offer significant improvement, even though it does not appear to penalize the performance either.
 \item redundancy removal performs better that novelty detection.
\end{itemize}

The experimental results presented judged only one aspect of our system; namely its responsiveness. Based on these results we have seen that combining different methods for the components of the overall summarization method, one can achieve significantly different results. It is very important, however, that the proposed summarization components offered competitive results through simple application of n-gram graph theory and methodologies, without any optimization on specific corpora.

As a side result, we have shown that there is a way to detect and remove noisy patterns from within n-gram graphs, using simple graph operators. Furthermore, we have illustrated that the removal of these patterns can improve the results of certain tasks. These tasks, like classification and topic detection, should be investigated through the n-gram graph representation prism to determine the potential of this representation as a generic NLP tool.

It is obvious from the experiments that individual components are not easy to judge as parts of a summarization system. Focused and exhaustive performance evaluations should be carried out to identify the impact of each component to the overall performance. It might also be needed that components are examined outside the summarization context, as stand-alone methods for chunking, semantic annotation, redundancy detection, etc.

In the future, we plan to test the effect of using various n-gram ranks within different parts of the summarization process. We have so far intuitively concluded that n-grams of lower ranks express the grammar model of a given language, \ie{} the set of allowed sequences of characters, while higher rank n-grams cross over the word boundaries and offer topic information. In~\cite{ggianna2008sse} there exists a methodology for the detection of statistically-determined important substrings of a text, called symbols within the context of that work. The use of these symbols, only, within n-gram graphs may alleviate the insertion of noise within the different summarization steps and diminish the computational cost of the method. Furthermore, emergent subgraphs and paths within a document set graph may allow for the extraction of non-obvious relations between text snippets as well as the detection of discourse phenomena and subtopics within a document set. These phenomena and subtopics can then be used to improve the structure and sentence ordering of the summary, as it has been shown that sentence ordering has an important effect on summarization~\cite{dang2009otSLIDES,barzilay2002iss}.


Last, but definitely not least, we need to evaluate the summaries extracted in any of the given corpora under the view of additional textual qualities --- \ie{} regardless of any responsiveness-related score. We should identify in what way the individual extracted summaries (see Figure \ref{ex:SummaryD0801A} for a sample summary and Appendix \ref{apd:Sample}) are worse from gold-standard summaries. Only then will we be able to improve on our promising current work.

We note that the whole \SYSTEM{} source code is available --- and under constant revision and improvement --- as part of the JINSECT open source project\footnote{See \url{http://sourceforge.net/projects/jinsect/} for more.} to facilitate the study of the methods we have presented.

\begin{figure}
\begin{verbatim}
 The newspaper International Herald Tribune reported on Friday  that production problems at one of Airbus's main parts plants in  Germany was at the root of the problem, rather than any safety or  quality issues.  Construction problems have delayed the introduction of the double-deck A380, the largest passenger plane in the world.  Prang said Airbus's management had made that announcement after analysing the production timetable for the whole project, and that  no one factor could be blamed for the delay.  We are in the process of reviewing the timetable. Airbus A380 superjumbo passes emergency evacuation test.
\end{verbatim}
\caption{\label{ex:SummaryD0801A}Sample summary for the D0801-A topic of TAC 2008 on the A380 airbus production and launch news.}
\end{figure}

\section*{acknowledgments}
The research described within this article was supported by the research and development project ONTOSUM\footnote{See also \url{http://www.ontosum.org/}}, which is in turn funded by the Greek General Secretariat for Research and Technology. 

As a sentence splitting module we have used the SentenceSplitter module of the JavaRAP toolkit (see \url{http://www.comp.nus.edu.sg/~qiul/NLPTools/JavaRAP.html}).

\bibliographystyle{apalike}
\bibliography{thesis}

\appendix
%
%
\section{Sample Summary and Scoring}\label{apd:Sample}
Here we provide a sample DUC 2006 corpus topic (Example \ref{ex:SampleSummaryTopic}) and the \SYSTEM{} extracted summary in Figure \ref{ex:SampleSummaryApndx}. 

\begin{example}\label{ex:SampleSummaryTopic}
\textbf{Topic Definition:}
\begin{verbatim}
<topic>
<num> D0601A </num>
<title> Native American Reservation System - pros and cons </title>

<narr>
Discuss conditions on American Indian 
reservations or among 
Native American communities. 
Include the benefits and drawbacks of the reservation system. 
Include legal privileges and problems.
</narr>
</topic> 
\end{verbatim}
\end{example}

\begin{figure}
\begin{verbatim}
Over all, the rate of substance abuse among Native American adults is over 20 percent nationwide.  A retail shopping center is proposed for a nearby section of the reservation.  As a result, some urban Native Americans feel driven away.  Juvenile crime is one strand in the web of social problems facing urban and reservation Indian communities, the report said.  He hopes to move his business to Union Gap, a few miles north and off the reservation.  Nearly 46,000 Native Americans live in the New York City metropolitan area.  Powless said the Onondaga people want to work with the community outside the reservation to improve the economy of the region, perhaps creating tourism destinations that might include Indian culture, or setting up a free-trade zone at unused manufacturing sites.  The Navajo Reservation is primarily dependent on federal money.  Reservations are home to some of the gravest poverty and worst health care in the country.  The issue was the tribe's operation of a shoe-box casino on its reservation.  Let's find a way to benefit from it.  ''      Bowing to a court ruling, Congress in 1988 laid out a statutory framework allowing American Indian tribes to offer high-stakes bingo games and casino-style gambling on their historic reservation lands.  The picture is similar on some smaller reservations, like the Fort Peck Reservation in eastern Montana.  But people in Neah Bay point out that Native American communities do not have a single, monolithic viewpoint any more than other communities.  
\end{verbatim}
\caption{Sample summary for the D0601A topic of DUC 2006}\label{ex:SampleSummaryApndx}
\end{figure}

In table \ref{tbl:scoringSample} we indicate the top 10 candidate sentences from the document set used to extract the summary. The sentences appear in decreasing order of their score. It is important to note that some of the sentences were removed from the \emph{redundancy removal step}, therefore not appearing in the final summary.

\begin{table}
 \scriptsize
 \begin{tabular}{r|p{0.8\textwidth}}
{\normalsize Value} &  {\normalsize Sentence} \\ \hline\hline
0.331996          &The income disparity on the reservation has widened.  \\ \hline
0.300312          &Many reservation residents are frustrated. \\  \hline
0.276393          &Over all, the rate of substance abuse among Native American adults is over 20 percent nationwide.  \\ \hline
0.269523          &But not everyone on the reservation is happy about the growth. \\ \hline 
0.256380          &A retail shopping center is proposed for a nearby section of the reservation. \\  \hline
0.255839          &And the migration from the reservations continues.  ``Mostly, it's economics,'' says Joanne Dunne, a spokeswoman for the Boston Indian Council, a nonprofit cultural group.  ``The reservation typically doesn't provide you with any real opportunities.  '' Also, many Native Americans travel between the reservation and urban areas.  ``Some tribes traditionally go from one place to another,'' says Dunne.  ``The Micmacs always crossed the border from Canada to come from time immemorial.  ''      In other cities, city and state governments have created agencies to specifically deal with the Native American population.  \\ \hline
0.252962          &As a result, some urban Native Americans feel driven away. \\ \hline 
0.235350          &From 1980 to 2000, the urban Native American population has more than doubled.  \\ \hline
0.235100          &But a non-Indian resident of a reservation has no say in tribal government. \\ \hline 
0.216477          &Smith and thousands like her are seeking help for their substance abuse at the American Indian Community House, the largest of a handful of Native American cultural institutions in the New York area.  ``Where else can I go for help?  '' she asks.  ``Any place else, they don't understand you like they do here.  ''      Native Americans around the country are leaving reservations and relocating in urban areas at a dizzying rate.  \\ \hline
 \end{tabular}
 \caption{The top 10 sentences provided from \SYSTEM{} and their scores (sentence scoring).}\label{tbl:scoringSample}
\end{table}

\end{document}